\documentclass[10pt,twocolumn,letterpaper]{article}

\usepackage{cvpr}              % To produce the CAMERA-READY version
\usepackage{booktabs}    % 优雅的表格线
\usepackage{multirow}    % 支持跨行合并
\usepackage{tabularx}    % 可选：自适应列宽
\usepackage{makecell}    % 可选：单元格内换行
\usepackage{hyperref}    % 超链接与引用跳转（通常已存在）
\usepackage{algorithm}
\usepackage{algpseudocode}
\usepackage[table]{xcolor}
\usepackage{graphicx}
\usepackage{float}
\usepackage{pifont}

\definecolor{cvprblue}{rgb}{0.21,0.49,0.74}
\def\paperID{} % *** Enter the Paper ID here
\def\confName{CVPR}
\def\confYear{2026}

\title{Iterative Inference-time Scaling with Adaptive Frequency Steering\\ for Image Super-Resolution}

\author{
    Hexin Zhang\textsuperscript{1},
    Dong Li\textsuperscript{1,\ding{41}},
    Jie Huang\textsuperscript{1},
    Bingzhou Wang\textsuperscript{1},
    Xueyang Fu\textsuperscript{1,\ding{41}},
    Zhengjun Zha\textsuperscript{1}
    \\
    \textsuperscript{1}University of Science and Technology of China\\
    {\tt\small zhanghexin@mail.ustc.edu.cn,dongli6@mail.ustc.edu.cn}
}

\begin{document}
\maketitle
\begin{abstract}
    \begingroup
    \renewcommand\thefootnote{} 
    \footnotetext{\ding{41} Corresponding authors.}
    \endgroup
Diffusion models have become a leading paradigm for image super-resolution (SR), but existing methods struggle to guarantee both the high-frequency perceptual quality and the low-frequency structural fidelity of generated images. Although inference-time scaling can theoretically improve this trade-off by allocating more computation, existing strategies remain suboptimal: reward-driven particle optimization often causes perceptual over-smoothing, while optimal-path search tends to lose structural consistency. To overcome these difficulties, we propose \textbf{Iterative Diffusion Inference-Time Scaling with Adaptive Frequency Steering (IAFS)}, a training-free framework that jointly leverages iterative refinement and frequency-aware particle fusion. 
IAFS addresses the challenge of balancing perceptual quality and structural fidelity by progressively refining the generated image through iterative correction of structural deviations. Simultaneously, it ensures effective frequency fusion by adaptively integrating high-frequency perceptual cues with low-frequency structural information, allowing for a more accurate and balanced reconstruction across different image details. 
Extensive experiments across multiple diffusion-based SR models show that IAFS effectively resolves the perception–fidelity conflict, yielding consistently improved perceptual detail and structural accuracy, and outperforming existing inference-time scaling methods.
\end{abstract}    
\begin{figure}[t]
    \centering 
    \includegraphics[width=1.0\linewidth]{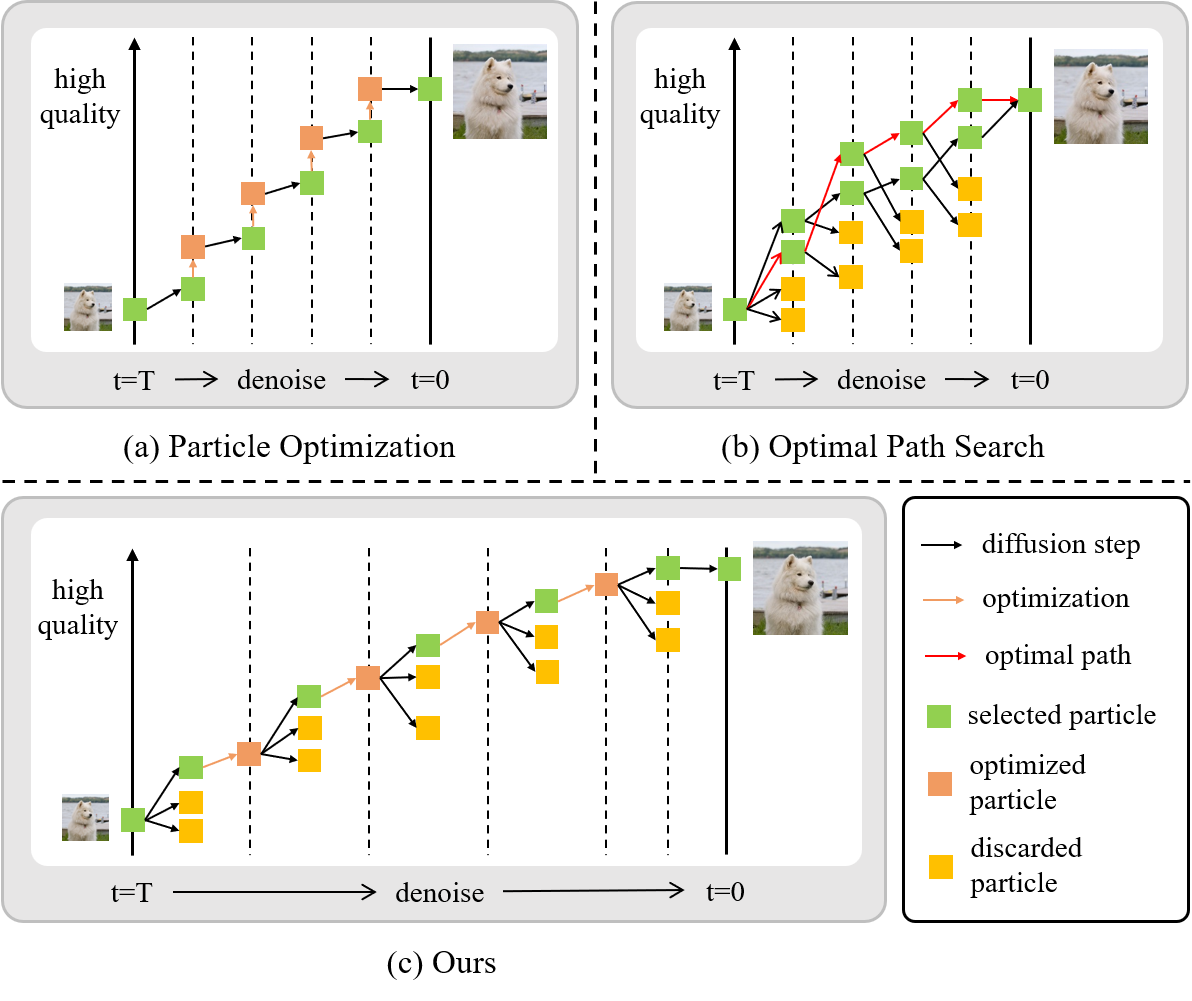} 
    \caption{Comparison of different inference-time scaling strategies. (a) Diffusion inference-time scaling based on \textbf{Particle Optimization}. (b) Diffusion inference-time scaling based on \textbf{Optimal Path Search}. (c) \textbf{Iterative} Diffusion inference-time scaling based on \textbf{Adaptive Frequency Steering (IAFS)}. } % 图片标题
    \label{fig:method} % 标签
\end{figure}

\section{Introduction}
\label{sec:intro}
Image super-resolution (SR) aims to recover a high-resolution (HR) image from its low-resolution (LR) counterpart.
Previous end-to-end SR models directly learns the mapping process from LR to HR spaces through hand-crafted priors~\cite{xia2015learning, yang2010image} or deep networks~\cite{glasner2009super, dong2015image}, producing the super-resolved (SR) images with high structural fidelity but leaving huge improvement space for perceptual quality.
More recently, diffusion models (DMs)~\cite{song2019generative, song2020denoising, kingma2021variational,fang2024vivid} exhibit powerful generative capabilities in modeling data distribution, which are effective in SR tasks  that improve the perceptual quality of SR images with vivid fine-grained details~\cite{rombach2022high, saharia2022image, li2022srdiff, yue2023resshift, lin2024diffbir, wu2024seesr, wu2024one, yang2024pixel}.

However, diffusion models struggle to balance perceptual quality and structural fidelity in the SR task~\cite{blau2018perception, saharia2022image}. 
Increasing the model's parameter size or computational budget can push the overall Pareto front of fidelity–perceptual quality towards a better solution domain~\cite{liang2024scaling}.
Therefore, we hope to use inference-time scaling to increase computational power or iteration in the inference process to improve this problem.
But directly applying native inference scaling to super-resolution (SR) is difficult because of image SR tasks differing from text-to-image tasks. On the one hand, the structural conditions of SR are stronger; on the other hand, the improvement in perceptual quality is limited because of the existing ground-truth images. This means that super-resolution has a higher requirement for balancing low-frequency fidelity and high-frequency perception. However, the core idea of existing inference scaling methods still relies on rewards, which leads to results biased towards reward models, making direct application to SR difficult.
 on the other hand, optimizing perceptual quality can easily produce illusions and deviate from the true reference image ~\cite{ledig2017photo}. Therefore, researchers proposed a training-free inference-time scaling technique to overcome this difficulty ~\cite{ma2025inference, uehara2025inference, singhal2025general, li2025dynamic, zhang2025inference, jajal2025inference, mao2025ctrl, hu2025kernel, farahbakhsh2025inference}, which improves the generated image quality with limited additional inference computation without changing the model weights. 

 We reviewed existing inference scaling methods. The mainstream methods that achieve good fidelity and perceptual quality balance can be divided into two categories: one directly optimizes multiple sampled trajectories (particles) at each timestep~\cite{mao2025ctrl, hu2025kernel, zhang2025inference, jajal2025inference}, as shown in Fig.~\ref{fig:method} (a); and the other directly finds the optimal path from multiple sampled trajectories generated during the inference process~\cite{singhal2025general, li2025dynamic, dang2025inference}, as shown in Fig.~\ref{fig:method} (b). However, as can be seen from Tab.~\ref{tab:1}, the particle optimization method often results in poor perceptual quality of the generated image, while the method of searching for the optimal path results in poor structural fidelity of the generated image.
 Furthermore, we find that reward models can guide the generation of high-frequency information in images to achieve good perceptual results, but low-frequency structures suffer from defects, while the low-frequency information of different candidate particles is complementary, and their combination helps improve fidelity.
 Therefore, our goal is to effectively integrate these two components so as to improve the capability of inference scaling to achieve a more favorable balance between fidelity and perceptual quality in SR.

To achieve this target, we propose Iterative Diffusion Inference-Time Scaling with Adaptive Frequency Steering (IAFS), as illustrated in Fig.~\ref{fig:method}(c). IAFS is built on two principles: iterative refinement and adaptive frequency steering. In SR tasks, high-frequency details are progressively recovered during diffusion, whereas low-frequency structures are largely determined in early steps, as proved in Sec.~\ref{sec:Preliminary}. Motivated by this observation, we adopt an iterative strategy that uses the output of each round as pseudo-GT to guide subsequent sampling, and introduce frequency-domain fusion at each timestep to adaptively combine high-frequency perceptual cues with low-frequency structural information.
In practice, we first select the best-performing particles at each timestep via a reward function, then decompose them into high-frequency and low-frequency components. The low-frequency part is aligned with similar low-frequency information from the reference particle pool to obtain an adaptive structural representation, which is fused with the high-frequency component to produce the optimal particle for the next step. During the iteration process, the first iteration uses CLIPIQA~\cite{wang2023exploring} to guide sampling, while subsequent iterations adopt a hybrid reward combining CLIPIQA and LPIPS~\cite{zhang2018unreasonable} to gradually approach a perceptual–fidelity compromise. Through iterative refinement and frequency fusion, IAFS effectively leverages low-frequency complementarity and high-frequency details across timesteps, yielding high-quality SR results.

The main contributions of this paper are as follows:
\begin{itemize}
    \item We propose an iterative hybrid reward–guided inference-time scaling method that improves the balance between perceptual quality and structural fidelity in diffusion-based super-resolution. By reformulating inference optimization as a search for optimal frequency information, our approach effectively guides sampling and enhances overall image quality.
    \item We design an Adaptive Frequency Steering (AFS) method to optimize the iterative sampling process. By aligning and fusing the high-frequency information of the optimal particle with the low-frequency components of high-similarity particles in the pool, AFS yields samples with both high perceptual quality and structural accuracy.
    \item We conducted extensive experiments to validate our proposed method in conjunction with SR models of different diffusions, demonstrating that our method can significantly improve the perceptual quality and structural fidelity of generated images, and achieve state-of-the-art results for image SR tasks compared with other diffusion inference-time scaling methods.
\end{itemize}

\section{Related Works}
\label{sec:formatting}
%-------------------------------------------------------------------------
\subsection{Diffusion Models for Image Super-Resolution}
Early explorations of diffusion models in the field of image super-resolution began with SRDiff~\cite{li2022srdiff} and SR3~\cite{saharia2022image}, which generated high-resolution images by training a diffusion model with low-resolution images as conditional inputs. However, this method has a high sampling cost. To address this problem, LDM-SR~\cite{rombach2022high} trains an autoencoder in a low-dimensional latent space and performs a diffusion process, which significantly reduces computational overhead. ResShift~\cite{yue2023resshift} reconstructs the forward and reverse Markov chains of traditional diffusion models by introducing a “residual shifting” process between low-resolution (LR) and high-resolution (HR) images, which greatly reduces the difficulty of model estimation. DiffBIR~\cite{lin2024diffbir} decouples the image restoration process into degradation Removal and information Regeneration, generating high-quality images by introducing a restoration model in the early stages. SeeSR~\cite{wu2024seesr} trains a Degradation-Aware Prompt Extractor (DAPE) to generate accurate semantic prompts, which in turn guide a pre-trained T2I model to produce detail-rich and semantically correct results even when handling semantically ambiguous low-resolution images. Despite these advancements, most diffusion-based SR frameworks can not fundamentally address the inference-time limitations that hinder the joint optimization of perceptual quality and structural fidelity.
%-------------------------------------------------------------------------
\subsection{Inference-Time Scaling for Diffusion Models}
Inference-time scaling aims to align pre-trained diffusion models with specific task objectives or quality metrics during generation without requiring additional training \cite{ma2025inference, uehara2025inference}. Current methods for the inference sampling process can be divided into particle optimization-based methods and optimal path search-based methods.
particle optimization-based methods guide particles to regions with stronger preferences by imposing constraints on them. Classical Search \cite{zhang2025inference} performs a global search during sampling while using gradients and reward scores to guide samples toward higher-reward local regions. Evolutionary algorithms-based method \cite{jajal2025inference} regards the diffusion model as a black box, optimizing the initial noise in the latent space through genetic algorithms and evolutionary strategies.
Optimal path search methods include beam search \cite{steinbiss1994improvements, li2025dynamic, oshima2025inference, guo2025training} and SMC (sequential monte carlo)-based \cite{singhal2025general, su2025navigating, dang2025inference} methods. Best-of-N (BON) \cite{valentini2017best, ma2025inference} as the simplest path search method, can be regarded as a simplified form of beam search. It achieves optimal path search by simultaneously maintaining N sampled trajectories and selecting the best output at the last timestep. Diffusion Latent Beam Search (DLBS) \cite{oshima2025inference} first proposed an Inference-time Scaling method based on beam search. At each timestep, B beams are initialized and K candidates are generated for each beam. Then, a reward function is used to prune the sampling path. DSearch \cite{li2025dynamic} considers that particles are not easy to evaluate in the first few time steps of the inference process. It proposes a dynamic beam search method, which realizes a dynamic search strategy by gradually increasing the beam width with each timestep. FK-steering method \cite{singhal2025general}introduces SMC into inference-time scaling and shows that earlier approaches, such as Practical SMC Guidance \cite{dang2025inference}, can be regarded as specific instances of it. The F-SMC method \cite{su2025navigating} introduces a temperature parameter that varies with the time step to adjust the reward during the FK resampling process.
%-------------------------------------------------------------------------
\subsection{Inference-time Scaling for Image SR}
In order to ensure that the generated images have both high perceptual quality and structural fidelity, inference-time scaling methods that do not require training have been gradually introduced to solve the Image SR problem. Kernel density steering (KDS)~\cite{hu2025kernel} as a particle optimization-based method, employs kernel density estimation (KDE)~\cite{davis2011remarks, parzen1962estimation} to locate the highest-density region of particle distributions at each inference step, and computes an analytical KDE gradient via mean shift algorithm~\cite{1000236}, then uses this gradient to guide all particles toward a high-density mode. To address the computational challenge of KDE in high-dimensional latent space, 
KDS introduces a patch-wise mechanism for patch-wise processing. But the method remains computationally expensive even with this design. Moreover, its ensemble-consensus behavior tends to bias toward the distribution mean in SR tasks, resulting in overly smooth reconstructions and degraded perceptual quality because it sacrifices high-frequency texture details for improved low-frequency fidelity.
The side-information approach~\cite{farahbakhsh2025inference} is an optimal path search-based method, which introduces auxiliary information during inference to estimate the expected reward of each particle, pruning suboptimal candidates at every denoising step. By constraining the search space with deterministic cues derived from reference images, it effectively mitigates semantically inconsistent reconstructions. However, this method relies heavily on the availability of side information, requiring auxiliary images similar to the target within the dataset, which greatly limits its applicability. In conclusion, inference-time scaling methods for image SR tasks should combine particle optimization strategy with optimal path search strategy to ensure the diversity of the sampling process and the perceptual-structural quality of the sampling results.

%-------------------------------------------------------------------------

\begin{figure*}[t!]
    \centering 
    \includegraphics[width=1.0\linewidth]{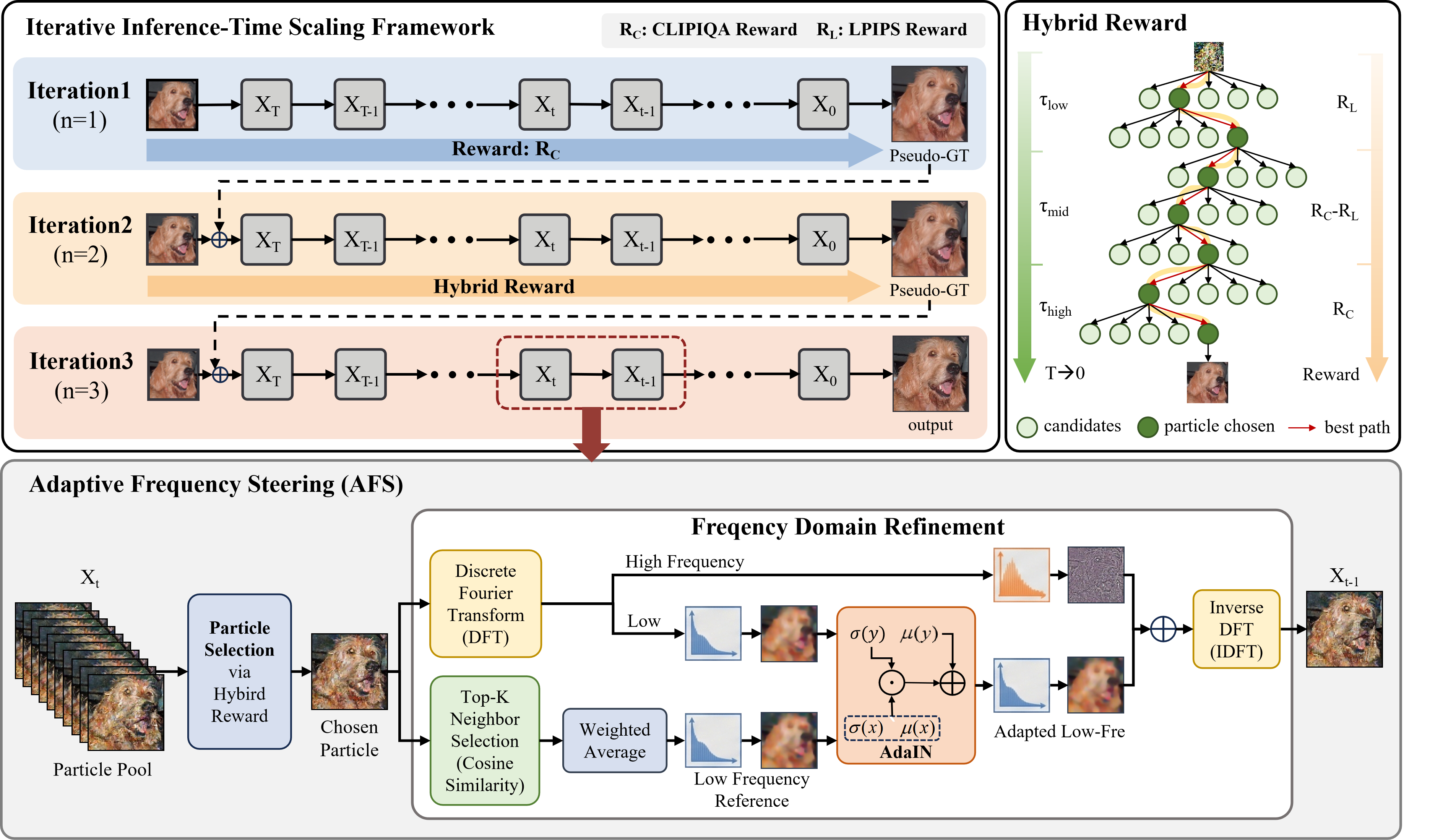} 
    \caption{The overall architecture of Iterative Inference-Time Scaling with Adaptive Frequency Steering. At each iteration, the diffusion output serves as pseudo-GT for the next one. For each timestep, $N$ particles are sampled and the chosen particle is selected via hybrid reward $R$. The top-$K$ most similar particles are averaged as the reference. The chosen particle is then decomposed into high/low-frequency components, aligned via Adaptive Instance Normalization (AdaIN), and fused to produce the optimal particle for the next timestep. } % 图片标题
    \label{fig:method} % 标签
\end{figure*}

\section{Methodology}

\subsection{Preliminary}
\label{sec:Preliminary}
\noindent\textbf{Reverse Process of Diffusion Models.}
The entire process of image generation using DMs is divided into the forward noise-adding stage and the reverse noise-denoising stage. When using pre-trained DMs for image generation, only the reverse denoising process is executed, since the forward process is fixed and implicitly captured by the model training procedure. Let the image corresponding to timestep $t$ be $\mathbf{x}_{t}$, and the time step corresponding to timestep $t-1$ be $\mathbf{x}_{t-1}$. Then the conditional probability distribution of $\mathbf{x}_{t-1}$ is formulated in Eq.~\ref{eq:posterior}, and can be equivalently reformulated into the sampling expression shown in Eq.~\ref{2}.
\begin{equation}
p_{\theta}(\mathbf{x}_{t-1} \mid \mathbf{x}_{t}) = \mathcal{N}\big(\mu_{\theta}(\mathbf{x}_{t}, t), \sigma_{t}^{2} \mathbf{I}\big).
\label{eq:posterior}
\end{equation}
\begin{equation}
\mathbf{x}_{t-1} = \mu_{\theta}(\mathbf{x}_{t}, t) + \sigma_{t} \mathbf{z}, \quad \mathbf{z} \sim \mathcal{N}(0, \mathbf{I}),
\label{2}
\end{equation}
where $\mu_{\theta}(\mathbf{x}_{t}, t)$ is the mean of the corresponding distribution, which is calculated by the pre-trained UNet network, and $\sigma_{t}$ is the variance.

\noindent\textbf{Inference-Time Scaling of Diffusion Models.}
Assume the optimal distribution for $\mathbf{x}_{t-1}$ during the backward reasoning process is $\mathbf{x}_{t-1}^*$. From Eq.~\ref{2}, we know that $\mu_{\theta}(\mathbf{x}_{t}, t)$ and $\sigma_{t}$ calculated by the pre-trained model are fixed values during inference. Therefore, the process of finding the optimal distribution is the same as the process of finding the optimal noise $\mathbf{z}^*$, which can be summarized as finding noise $\mathbf{z}$ that maximizes the expected reward \textit{R(x)} of the sample $\mathbf{x}_{t-1}$ from the pre-trained diffusion model $\mathbf{p_{\theta}}$, as shown in Eq.~\ref{3}.
\begin{equation}
z^{\star} = \arg\max_{z \in I} \mathbb{E}_{x \sim p_{\theta}(x \mid \psi)} \left[ R(x) \right].
\label{3}
\end{equation}

\noindent\textbf{Frequency Representation of Reverse Process.}
Finding the optimal noise using evaluation methods during the inference process of diffusion is very difficult, but the optimal noise can be obtained through frequency domain decomposition, thereby obtaining the optimal particles. 
Firstly, we perform a Fourier transform on Eq.~\ref{2} to obtain the spectrum corresponding to the distribution of $\mathbf{x}_{t-1}$ as follows:
\begin{equation}
\hat{f}_{\mathbf{x}_{t-1}}(\boldsymbol{\omega}) = \exp\left( -i\,\boldsymbol{\omega}^{\top}\boldsymbol{\mu}_{\theta} - \frac{1}{2}\sigma_{t}^{2}\|\boldsymbol{\omega}\|^{2} \right)
\label{4}
\end{equation}
Since $\mu_{\theta}(\mathbf{x}_{t}, t)$ and $\sigma_{t}$ are fixed values obtained from the pre-trained model at each timestep in the inference process, the first term in Eq. (\ref{4}) is the phase term, and thus the amplitude spectrum can be expressed as follows:
\begin{equation}
|\hat{f}(\boldsymbol{\omega})| = e^{-\frac{1}{2}\sigma_{t}^{2}\|\boldsymbol{\omega}\|^{2}}.
\label{5}
\end{equation}
In diffusion models, the value of $\sigma_{t}$ is relatively large in the early stage of the inference process. Thus, $\hat{f}(\boldsymbol{\omega})$ decays rapidly to 0 as $\omega$ increases and the signal is closer to pure noise in the spatial domain. In the later stage of the inference process, the value of $\sigma_{t}$ decreases, high-frequency information is gradually recovered and more details are restored. Therefore, in the early stage of the inference phase, the optimal particle can be found by evaluating the low-to-mid-frequency information of $\mathbf{x}_{t-1}$, while in the later stage of the inference phase, the optimal particle can be found by evaluating the high-frequency information of $\mathbf{x}_{t-1}$.

\subsection{Iterative Inference-Time Scaling Method}
Research shows that the perception metric CLIPIQA \cite{wang2023exploring} is more sensitive to the high-frequency information of images; the CLIPIQA value increases as the high-frequency information of the image becomes more complete \cite{wu2025one}. In contrast, the LPIPS \cite{zhang2018unreasonable} metric favors the mid-to-low-frequency structure of images \cite{liao2022deepwsd}. These complementary frequency sensitivities suggest that CLIPIQA and LPIPS can be jointly utilized as a reward function to guide the diffusion inference process. 
Following this insight, we assign different roles to the two metrics across the denoising trajectory. In the early stage of the inference phase, LPIPS is used to evaluate the low-frequency information of the particles to find the best particle. In the middle stage of the inference phase, a combination of LPIPS and CLIPIQA is used to evaluate the frequency information of the particles. Finally, in the later stage of the inference phase, CLIPIQA is used to evaluate the high-frequency information of the particles, ensuring that the best particles can be selected at each timestep.
However, using LPIPS throughout the entire inference process is not feasible because LPIPS is a reference-based metric requiring ground-truth images, whereas CLIPIQA is a no-reference metric and can reliably filter frequency information even without explicit supervision \cite{wu2025one}. To overcome this limitation, we design an iterative hybrid reward-guided strategy. In the first iteration, CLIPIQA is used as the reward for particle selection, and the generated result is used as the pseudo-gt for the next iteration. Then particle selection is performed by combining the pseudo-gt with the hybrid reward that changes with the inference process. And this process is repeated \textit{n} times, ultimately producing a result that balances both perceptual quality and structural fidelity. The pseudo-code of this process is shown in Algorithm~\ref{alg:iafs}. 

\begin{algorithm}[t]
\caption{Iterative Inference-Time Scaling}
\label{alg:iafs}
\begin{algorithmic}[1]
\Require Diffusion kernel $p_\theta(x_{t-1}\mid x_t)$; denoiser $\phi_t$; \\
         iterations $n$; particles $N$; \\
         frequency segments $\mathcal{T}_{\text{low}},\mathcal{T}_{\text{mid}},\mathcal{T}_{\text{high}}$
\State Initialize pseudo reference $\tilde{y}^{(0)} \gets \varnothing$

\For{$i = 1,\dots,n$}
    \State $y^{(i)} \gets 
        \begin{cases}
            \varnothing, & i = 1,\\
            \tilde{y}^{(i-1)}, & i > 1
        \end{cases}$

    \State Initialize particles $x_T^{(k)} \sim \mathcal{N}(0,I)$, $k=1,\dots,N$

    \For{$t = T,\dots,1$}
        \State Sample $x_{t-1}^{(k)} \sim p_\theta(x_{t-1}\mid x_t^{(k)})$ and predict $\hat{u}_t^{(k)} = \phi_t(x_t^{(k)})$
        \State Compute rewards $r_k$ for $k=1,\dots,N$ as
        \[
            r_k =
            \begin{cases}
                R_C(\hat{u}_t^{(k)}), & i = 1,\\[2pt]
                -R_L(\hat{u}_t^{(k)}, y^{(i)}), & i > 1,\, t \in \mathcal{T}_{\text{low}},\\[2pt]
                R_C(\hat{u}_t^{(k)}) - R_L(\hat{u}_t^{(k)}, y^{(i)}), & i > 1,\, t \in \mathcal{T}_{\text{mid}},\\[2pt]
                R_C(\hat{u}_t^{(k)}), & i > 1,\, t \in \mathcal{T}_{\text{high}} .
            \end{cases}
        \]
        \State Select $k^\star = \arg\max_k r_k$ and apply AFS to refine $x_t^{(k^\star)}$
    \EndFor

    \State Decode $\hat{x}^{(i)} = \phi_0(x_0^{(k^\star)})$ and set $\tilde{y}^{(i)} \gets \hat{x}^{(i)}$
\EndFor

\State \Return final image $\hat{x}^{(n)}$
\end{algorithmic}
\end{algorithm}

\subsection{Adaptive Frequency Steering}
Using only rewards to select particles during diffusion inference can lead to an over-preference for reward metrics in the generated results. This is especially true when selecting an optimal particle and constructing a particle pool at each timestep; if the actual quality of the particle is poor, the entire inference process can collapse. Therefore, it is crucial to optimize and guide the particles during the search process simultaneously. KDS~\cite{hu2025kernel} has demonstrated that the particles in the particle pool at each timestep possess "collective intelligence." Based on our analysis of frequencies during the inference process of diffusion, we designed an adaptive frequency steering strategy to optimize the optimal particle as shown in Fig.~\ref{fig:method}. Let the particle pool at timestep~\textit{t} be $\{ \mathbf{x}_{t}^{i} \}_{i=1}^{N}$
Firstly, we use a mixed reward system to select the best particle $\mathbf{x}_{t}^*$ as follows:
\begin{equation}
\mathbf{x}_{t}^{\star} = \arg\max_{i} R(\mathbf{x}_{t}^{i}),
\label{6}
\end{equation}
Then perform a Discrete Fourier Transform (DFT) on it to obtain the spectrum:
\begin{equation}
\begin{aligned}
\mathcal{F}(\mathbf{x}_{t}^{\star}) &= \mathbf{F}_{L}^{\star} + \mathbf{F}_{H}^{\star}, 
\label{7}
\end{aligned}
\end{equation}
\begin{equation}
\mathbf{F}_{L}^{\star} = \mathbf{M}_{L} \odot \mathcal{F}(\mathbf{x}_{t}^{\star}), \quad
\mathbf{F}_{H}^{\star} = \mathbf{M}_{H} \odot \mathcal{F}(\mathbf{x}_{t}^{\star}),
\label{8}
\end{equation}
where $\mathbf{M}_{L}$ and $\mathbf{M}_{H}$ are the low-frequency and high-frequency masks.
The remaining particles are selected from the top K similar neighbors based on cosine similarity \cite{xia2015learning} as follows:
\begin{equation}
\omega_{i} = \frac{\langle \mathbf{x}_{t}^{i}, \mathbf{x}_{t}^{\star} \rangle}{\|\mathbf{x}_{t}^{i}\| \, \|\mathbf{x}_{t}^{\star}\|}, \quad
\mathcal{N}_{K} = \text{Top-}K(\omega_{i}),
\label{9}
\end{equation}
where $\omega_{i}$ represents similarity weight for each particle with $\mathbf{x}_{t}^*$. And then a low-frequency reference is obtained by weighted averaging as follows:
\begin{equation}
\overline{\mathbf{F}}_{L} =
\frac{\sum_{\mathbf{x}_{t}^{(k)} \in \mathcal{N}_{K}} \omega_{k} \mathbf{M}_{L} \odot \mathcal{F}(\mathbf{x}_{t}^{(k)})}
{\sum_{\mathbf{x}_{t}^{(k)} \in \mathcal{N}_{K}} \omega_{k}}.
\label{10}
\end{equation}
where $\overline{\mathbf{F}}_{L}$ is a low-frequency reference, which aggregates low-frequency commonalities among multiple particles. 
To ensure style alignment between the master particle and the reference particle, we introduce Adaptive Instance Normalization (AdaIN) \cite{huang2017arbitrary} to compute an adaptive low-frequency spectrum as follows:
\begin{equation}
\hat{\mathbf{F}}_{L} = \sigma(\overline{\mathbf{F}}_{L}) 
\frac{\mathbf{F}_{L}^{\star} - \mu(\mathbf{F}_{L}^{\star})}{\sigma(\mathbf{F}_{L}^{\star})} 
+ \mu(\overline{\mathbf{F}}_{L}),
\label{11}
\end{equation}
where $\mu$ is the mean of channel and $\sigma$ is variance of channel.
Finally, the fused spectrum is obtained by adding the high-frequency component $\mathbf{F}_{H}^{\star}$ of the main particle to the adaptive low-frequency spectrum $\hat{\mathbf{F}}_{L}$.

\begin{table*}[t!]
\centering
\caption{Super-Resolution Performance compared with different inference-time scaling methods on ImageNet and DIV2K. The best and second best results are marked in \textcolor{red}{red} and \textcolor{blue}{blue}.}
\label{tab:1}
\setlength{\tabcolsep}{5pt}
\renewcommand{\arraystretch}{1.05}
\footnotesize
\resizebox{\textwidth}{!}{%
\begin{tabular}{ccccccc c cccccc}
\toprule
\multicolumn{1}{c}{Datasets} &
\multicolumn{6}{c}{ImageNet} &
\multicolumn{6}{c}{DIV2K} \\
\cmidrule(lr){2-7}\cmidrule(lr){8-13}
\multicolumn{1}{c}{Method} &
PSNR & SSIM & LPIPS($\downarrow$) & CLIP-IQA & MUSIQ & MANIQA &
PSNR & SSIM & LPIPS($\downarrow$) & CLIP-IQA & MUSIQ & MANIQA \\
\midrule
\textbf{ResShift} 
& 23.01 & 0.6466 & 0.2371 & 0.5860 & 53.18 & 0.4191
& 24.79 & 0.6273 & 0.3365 & 0.5418 & 59.52 & 0.4125 \\

+BON
& 23.31 & 0.6504 & 0.2108 & 0.7038 & 59.53 & 0.4938
& 25.01 & 0.6305 & 0.3182 & 0.6259 & 63.75 & 0.4922 \\

+BS
& 23.38 & 0.6469 & \textcolor{blue}{0.2079} & \textcolor{red}{0.7823} & \textcolor{red}{60.94} & \textcolor{red}{0.5026}
& 25.09 & 0.6281 & \textcolor{blue}{0.3126} & \textcolor{red}{0.6758} & \textcolor{red}{64.37} & \textcolor{red}{0.5039} \\

+SMC
& 23.56 & 0.6547 & 0.2224 & 0.6203 & 56.45 & 0.4339
& 25.14 & 0.6257 & 0.3285 & 0.5615 & 61.49 & 0.4197 \\

+KDS
& \textcolor{blue}{23.73} & \textcolor{blue}{0.6554} & 0.2379 & 0.5904 & 54.30 & 0.3902
& \textcolor{blue}{25.21} & \textcolor{blue}{0.6348} & 0.3367 & 0.5502 & 60.05 & 0.3955 \\

\rowcolor{gray!20}
\textbf{+IDAFS}
& \textcolor{red}{24.10} & \textcolor{red}{0.6558} & \textcolor{red}{0.2072} & \textcolor{blue}{0.7533} & \textcolor{blue}{59.83} & \textcolor{blue}{0.4954}
& \textcolor{red}{25.45} & \textcolor{red}{0.6377} & \textcolor{red}{0.3119} & \textcolor{blue}{0.6449} & \textcolor{blue}{63.92} & \textcolor{blue}{0.4972} \\
\specialrule{0.1pt}{0pt}{0pt}

\textbf{DiffBIR}
& 22.73 & 0.6189 & 0.2601 & 0.6635 & 65.26 & 0.5408
& 23.64 & 0.5528 & 0.3629 & 0.7734 & 67.31 & 0.5948 \\

+BON
& 22.80 & 0.6201 & 0.2620 & \textcolor{red}{0.7422} & 66.45 & 0.5643
& 23.46 & 0.5490 & 0.3511 & \textcolor{blue}{0.8452} & 69.14 & 0.6155 \\

+BS
& 23.41 & 0.6190 & \textcolor{blue}{0.2424} & 0.6963 & \textcolor{blue}{67.78} & \textcolor{red}{0.5779}
& 23.48 & 0.5526 & \textcolor{blue}{0.3492} & \textcolor{red}{0.8592} & \textcolor{red}{69.97} & \textcolor{red}{0.6221} \\

+SMC
& 23.74 & 0.6261 & 0.2530 & 0.6711 & 65.64 & 0.5549
& 23.75 & 0.5601 & 0.3665 & 0.7829 & 67.54 & 0.6083 \\

+KDS
& \textcolor{blue}{23.85} & \textcolor{red}{0.6297} & 0.2597 & 0.6658 & 65.33 & 0.5254
& \textcolor{blue}{23.79} & \textcolor{red}{0.5658} & 0.3587 & 0.7765 & 67.43 & 0.5732 \\

\rowcolor{gray!20}
\textbf{+IDAFS}
& \textcolor{red}{23.98} & \textcolor{blue}{0.6285} & \textcolor{red}{0.2403} & \textcolor{blue}{0.7033} & \textcolor{red}{67.89} & \textcolor{blue}{0.5722}
& \textcolor{red}{23.86} & \textcolor{blue}{0.5617} & \textcolor{red}{0.3475} & 0.7969 & \textcolor{blue}{69.25} & \textcolor{blue}{0.6204} \\

\specialrule{0.1pt}{0pt}{0pt}

\textbf{SeeSR}
& 23.10 & 0.6258 & 0.2343 & 0.6623 & 64.95 & 0.5172
& 23.61 & 0.5846 & 0.2983 & 0.7602 & 67.78 & 0.5533 \\

+BON
& 23.42 & 0.6331 & 0.2277 & \textcolor{red}{0.7819} & 68.81 & 0.5975
& 23.85 & \textcolor{blue}{0.5917} & 0.2956 & \textcolor{red}{0.8479} & 71.35 & \textcolor{red}{0.6318} \\

+BS
& 23.51 & 0.6267 & \textcolor{blue}{0.2223} & 0.7657 & \textcolor{red}{69.14} & \textcolor{red}{0.6021}
& 24.26 & 0.5859 & \textcolor{blue}{0.2887} & \textcolor{blue}{0.8301} & \textcolor{red}{71.64} & 0.6172 \\

+SMC
& 23.64 & \textcolor{blue}{0.6372} & 0.2305 & 0.6789 & 67.93 & 0.5314
& \textcolor{blue}{24.63} & 0.5815 & 0.2895 & 0.7723 & 70.32 & 0.5749 \\

+KDS
& \textcolor{blue}{23.81} & 0.6362 & 0.2373 & 0.6678 & 65.93 & 0.4922
& 23.79 & 0.5833 & 0.2932 & 0.7620 & 68.31 & 0.5372 \\

\rowcolor{gray!20}
\textbf{+IDAFS}
& \textcolor{red}{23.87} & \textcolor{red}{0.6432} & \textcolor{red}{0.2175} & \textcolor{blue}{0.7659} & \textcolor{blue}{68.90} & \textcolor{blue}{0.5984}
& \textcolor{red}{24.91} & \textcolor{red}{0.5925} & \textcolor{red}{0.2815} & 0.7836 & \textcolor{blue}{71.58} & \textcolor{blue}{0.6277} \\

\specialrule{0.1pt}{0pt}{0pt}

\end{tabular}
}
\end{table*}

\begin{figure*}[t!]
    \centering 
    \includegraphics[width=0.8\linewidth]{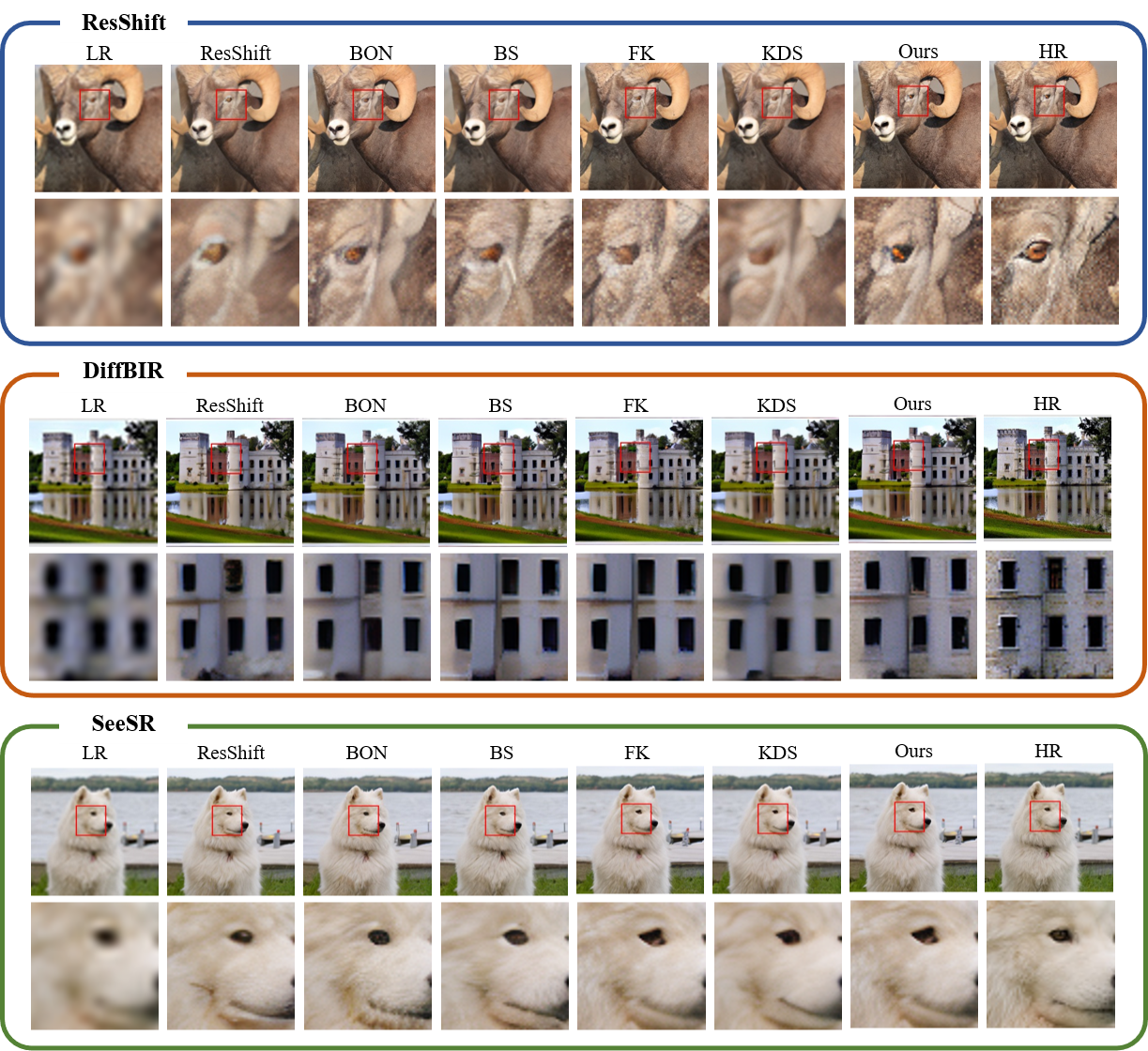} 
    \caption{Qualitative comparison of 4$\times$ Super-Resolution with different inference-time scaling methods on baselines.} % 图片标题
    \label{fig:experiment} % 标签
% \vspace{-0.5em}    
\end{figure*}

\begin{table*}[t!]
\centering
\caption{Performance comparison of inference-time scaling methods on RealSR and DRealSR datasets. The best and second best results are marked in \textcolor{red}{red} and \textcolor{blue}{blue}.}
\label{tab:2}
\setlength{\tabcolsep}{5pt}
\renewcommand{\arraystretch}{1.05}
\footnotesize
\resizebox{\textwidth}{!}{
\begin{tabular}{ccccccc c cccccc}
\toprule
\multirow{2}{*}{Method} &
\multicolumn{6}{c}{RealSR} &
\multicolumn{6}{c}{DRealSR} \\
\cmidrule(lr){2-7}\cmidrule(lr){8-13}
\multicolumn{1}{c}{Method} &
PSNR & SSIM & LPIPS($\downarrow$) & CLIP-IQA & MUSIQ & MANIQA &
PSNR & SSIM & LPIPS($\downarrow$) & CLIP-IQA & MUSIQ & MANIQA \\
\specialrule{0.1pt}{0pt}{0pt}

\textbf{ResShift}
& 24.32 & 0.6230 & 0.2860 & 0.5590 & 58.69 & 0.4593
& 26.42 & 0.6431 & 0.3416 & 0.5248 & 56.81 & 0.4423 \\

+BON
& 24.58 & 0.6353 & 0.2853 & 0.5814 & \textcolor{blue}{64.78} & \textcolor{blue}{0.5325}
& 26.54 & 0.6505 & 0.3265 & 0.5800 & 61.23 & \textcolor{blue}{0.5058} \\

+BS
& 24.65 & 0.6370 & \textcolor{blue}{0.2785} & \textcolor{red}{0.6110} & \textcolor{red}{66.15} & \textcolor{red}{0.5530}
& 26.63 & 0.6520 & \textcolor{blue}{0.3180} & \textcolor{red}{0.5937} & \textcolor{red}{63.04} & \textcolor{red}{0.5257} \\

+SMC
& 24.72 & 0.6323 & 0.2830 & 0.5782 & 61.27 & 0.4991
& 26.68 & \textcolor{blue}{0.6531} & 0.3374 & 0.5571 & 58.90 & 0.4672 \\

+KDS
& \textcolor{blue}{24.86} & \textcolor{red}{0.6420} & 0.2865 & 0.5634 & 59.51 & 0.4557
& \textcolor{blue}{26.71} & \textcolor{red}{0.6555} & 0.3389 & 0.5462 & 57.79 & 0.4355 \\

\rowcolor{gray!20}
\textbf{+IDAFS}
& \textcolor{red}{24.91} & \textcolor{blue}{0.6395} & \textcolor{red}{0.2768} & \textcolor{blue}{0.5920} & 64.25 & 0.5231
& \textcolor{red}{26.75} & 0.6539 & \textcolor{red}{0.3150} & \textcolor{blue}{0.5871} & \textcolor{blue}{61.82} & 0.4971 \\

\specialrule{0.1pt}{0pt}{0pt}
\textbf{DiffBIR}
& 24.70 & 0.6185 & 0.3035 & 0.6178 & 65.78 & 0.6221
& 24.57 & 0.6581 & 0.3648 & 0.6372 & 64.95 & 0.6052 \\

+BON
& 24.85 & 0.6245 & 0.3005 & 0.6472 & 67.25 & 0.6441
& 24.72 & 0.6639 & 0.3521 & 0.6657 & 67.16 & 0.6379 \\

+BS
& 25.03 & 0.6279 & \textcolor{blue}{0.2945} & \textcolor{red}{0.6623} & \textcolor{red}{68.31} & \textcolor{red}{0.6572}
& 24.80 & 0.6651 & \textcolor{blue}{0.3445} & \textcolor{red}{0.6824} & \textcolor{red}{68.33} & \textcolor{red}{0.6482} \\

+SMC
& 25.10 & 0.6202 & 0.2978 & \textcolor{blue}{0.6571} & 66.15 & 0.6453
& \textcolor{blue}{24.81} & \textcolor{blue}{0.6705} & 0.3501 & 0.6705 & 66.27 & 0.6211 \\

+KDS
& \textcolor{blue}{25.15} & \textcolor{red}{0.6331} & 0.3010 & 0.6348 & 65.82 & 0.6110
& 24.76 & 0.6520 & 0.3526 & 0.6520 & 65.39 & 0.6035 \\

\rowcolor{gray!20}
\textbf{+IDAFS}
& \textcolor{red}{25.21} & \textcolor{blue}{0.6314} & \textcolor{red}{0.2925} & 0.6552 & \textcolor{blue}{67.39} & \textcolor{blue}{0.6503}
& \textcolor{red}{24.95} & \textcolor{red}{0.6751} & \textcolor{red}{0.3417} & \textcolor{blue}{0.6751} & \textcolor{blue}{67.94} & \textcolor{blue}{0.6473} \\

\specialrule{0.1pt}{0pt}{0pt}
\textbf{SeeSR}
& 25.38 & 0.6016 & 0.2949 & 0.6612 & 65.77 & 0.5882
& 25.27 & 0.6216 & 0.3168 & 0.6342 & 65.03 & 0.5894 \\

+BON
& 25.45 & 0.6072 & 0.2918 & \textcolor{blue}{0.6874} & 67.25 & 0.6175
& 25.39 & 0.6265 & 0.3054 & \textcolor{blue}{0.6620} & 67.41 & 0.6257 \\

+BS
& 25.50 & 0.6081 & \textcolor{blue}{0.2880} & \textcolor{red}{0.6923} & \textcolor{red}{68.17} & \textcolor{red}{0.6359}
& 25.44 & 0.6281 & \textcolor{blue}{0.2976} & 0.6553 & \textcolor{red}{68.13} & \textcolor{red}{0.6482} \\

+SMC
& 25.58 & 0.6109 & 0.2905 & 0.6745 & 66.59 & 0.6027
& 25.52 & 0.6305 & 0.3037 & 0.6510 & 66.29 & 0.6154 \\

+KDS
& \textcolor{blue}{25.65} & \textcolor{red}{0.6148} & 0.2940 & 0.6627 & 65.25 & 0.5811
& \textcolor{blue}{25.53} & \textcolor{red}{0.6327} & 0.3125 & 0.6489 & 65.04 & 0.5855 \\

\rowcolor{gray!20}
\textbf{+IDAFS}
& \textcolor{red}{25.70} & \textcolor{blue}{0.6115} & \textcolor{red}{0.2865} & 0.6800 & \textcolor{blue}{67.80} & \textcolor{blue}{0.6257}
& \textcolor{red}{25.66} & \textcolor{blue}{0.6310} & \textcolor{red}{0.2956} & \textcolor{red}{0.6621} & \textcolor{blue}{67.50} & \textcolor{blue}{0.6315} \\

\specialrule{0.1pt}{0pt}{0pt}
\end{tabular}
}
% \vspace{-0.5em}
\end{table*}

\section{Experiments}
\subsection{Experiment Setting}
We select ImageNet-Test~\cite{deng2009imagenet, yue2023resshift} that contains 3,000 images and DIV2K-val~\cite{agustsson2017ntire} as datasets for synthetic image SR evaluation. RealSR~\cite{cai2019toward} and DRealSR~\cite{wei2020component} are adopted for real-world image SR evaluation. We demonstrate the plug-and-play nature of our algorithms by considering three different baselines: ResShift~\cite{yue2023resshift}, DiffBIR~\cite{lin2024diffbir}, and SeeSR~\cite{wu2024seesr}. For the three baselines, we set their inference timesteps to 15, 50, and 50 during the experiment. To demonstrate the effectiveness of our method, we compared it with Best-of-N (BON)~\cite{ma2025inference}, Beam Search (BS)~\cite{li2025dynamic}, FK-steering (FK)~\cite{singhal2025general}, and Kernel Density Steering (KDS)~\cite{hu2025kernel}. To ensure the computational budget during inference is equivalent across different methods, \textit{N} equal 10 sampling paths are maintained for BON, FK, and KDS during the sampling process, and BS selects the best beam by sampling \textit{B} equal 10 beams at each timestep, same as our method. Furthermore, the peripheral reward functions for BON, BS, and FK use the same hybrid reward configuration as our method, and we also perform n=3 iterations for each method.

\subsection{Evaluation on Image Super-Resolution}
To evaluate the effectiveness of our method, we compare IAFS against existing inference-time scaling methods across three diffusion SR baselines—ResShift, DiffBIR, and SeeSR. Quantitative comparisons on ImageNet-Test and DIV2K-val are presented in Tab.~\ref{tab:1}, and results on RealSR and DRealSR are shown in Tab.~\ref{tab:2}.
Across all baselines, IAFS consistently yields substantial improvements over the original models. For each backbone, our method enhances both structural fidelity (PSNR) and structural consistency (SSIM), while simultaneously achieving superior perceptual quality reflected in LPIPS, CLIPIQA, MUSIQ, and MANIQA scores. These results demonstrate that IAFS not only strengthens low-frequency structural reconstruction but also recovers richer high-frequency details, thereby achieving a more balanced perceptual–structural trade-off than the vanilla baselines.
When compared with other inference-time scaling strategies, the advantages of IAFS become more explicit. BON and BS introduce only mild guidance during sampling, resulting in limited perceptual improvement and often failing to meaningfully enhance image detail. SMC methods provide stronger particle exploration but still struggle to maintain structural consistency, leading to unstable improvements across metrics. KDS, as a kernel-density-based particle optimization method, tends to bias all trajectories toward similar high-density modes, which improves PSNR/SSIM in some cases but often suppresses high-frequency textures, producing visually smooth yet perceptually degraded outputs. In contrast, IAFS consistently outperforms these methods by simultaneously improving perceptual metrics and maintaining or surpassing structural fidelity, demonstrating clear robustness across different diffusion SR frameworks.
Qualitative results in Fig.~\ref{fig:experiment} further validate these findings. Compared with other methods, our approach better restores fine-grained textures, produces fewer artifacts, and preserves sharper edges while avoiding the perceptual over-enhancement or structural collapse observed in competing methods. Most importantly, IAFS effectively leverages complementary frequency information to avoid the two common failure modes in SR diffusion inference—perceptually strong but structurally inaccurate outputs, and structurally faithful but perceptually dull outputs.

\subsection{Impact of Hyperparameters}
As an iterative guidance strategy, the performance of IAFS is closely related to its key hyperparameters: the number of sampled particles \textit{N} and the number of reference particles \textit{K} at each timestep as well as the number of iterations \textit{n}.
We analyze their impact on SR performance (LPIPS, PSNR, SSIM, CLIPIQA, MUSIQ, MANIQA) using the ResShift model on the ImageNet dataset.

\begin{figure}[t]
    \centering 
    \includegraphics[width=0.8\linewidth]{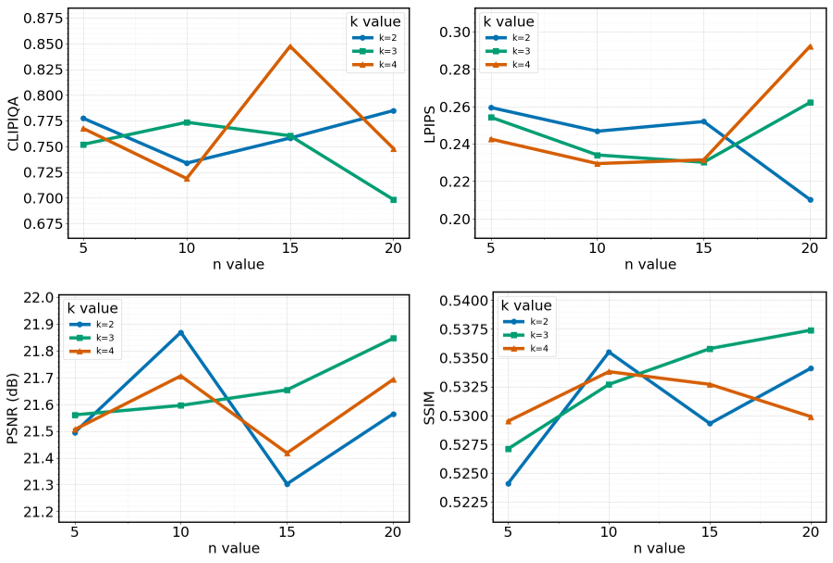} 
    \caption{Influence of the number of sampled particles ($N$) and the number of reference particles ($K$) on super-resolution performance. The perceptual metrics (CLIPIQA and LPIPS) and structural metrics (PSNR and SSIM) are plotted against different values of $N$ under three reference-particle settings ($K=2, 3, 4$).}% 图片标题
    \label{fig:plot} % 标签
\end{figure}

\noindent\textbf{Number of Samples \textit{N} and Reference Particles \textit{K}.}
The number of particles sampled at each timestep (\(N\)) determines the diversity of sampling paths and directly affects optimal-path selection, while the number of reference particles (\(K\)) influences the effectiveness of Adaptive Frequency Steering (AFS). As shown in Fig.~\ref{fig:plot}, both perceptual metrics (LPIPS, CLIPIQA) and structural metrics (PSNR, SSIM) exhibit consistent trends as \(N\) and \(K\) vary. Increasing \(N\) from 5 to 10 yields notable gains across all four metrics, but further increasing \(N\) to 20 produces only marginal improvements, indicating saturation relative to the rising computational cost. For \(K\), the results consistently show that \(K=2\) outperforms \(K=3\) and \(K=4\) in all metrics, while larger \(K\) values provide no additional benefit and often degrade performance. Considering both accuracy and efficiency, \(N=10\) captures most of the achievable improvement, and \(K=2\) offers the most reliable frequency guidance. Therefore, the combination \(N=10\) and \(K=2\) provides the optimal balance between perceptual quality, structural fidelity, and computation.

\noindent\textbf{Number of Iterations \textit{n}.}
The number of iterations plays a crucial role in balancing the perception distortion trade-off and determining computational cost.
Tab.~\ref{tab:imr_results} shows that iterative refinement improves perceptual metrics such as LPIPS and CLIPIQA in the early stages, with noticeable gains from $n=1$ to $n=3$. Beyond the third iteration, the improvement rate diminishes sharply, and by the seventh iteration the perceptual indices essentially reach saturation. A similar trend is observed in structural fidelity: PSNR and SSIM increase steadily during the first few iterations, but their gains plateau as the iteration count continues to rise. Notably, the marginal benefits obtained from iterations beyond $n=3$ are minimal. Therefore, considering both efficiency and the perception--structure balance, selecting a moderate iteration count of $n=3$ provides the most favorable trade-off, delivering stable improvements on both perceptual quality and structural fidelity while avoiding redundant computation.

\begin{table}[t]
\centering
\caption{Performance of IAFS on ResShift across different iterations.}
\label{tab:imr_results}
\resizebox{\columnwidth}{!}{
\begin{tabular}{c|cccccc}
\toprule
\text{iteration} & LPIPS($\downarrow$) & PSNR & SSIM & CLIPIQA & MUSIQ & MANIQA \\ 
\midrule
$n=1$ & 0.2105 & 24.05 & 0.6450 & 0.7504 & 59.71 & 0.4837 \\
$n=2$ & 0.2093 & 24.07 & 0.6453 & 0.7527 & 59.80 & 0.4841 \\
$n=3$ & 0.2082 & 24.10 & 0.6458 & 0.7533 & 59.83 & 0.4854 \\
$n=4$ & 0.2084 & 24.09 & 0.6457 & 0.7540 & \textbf{59.87} & \textbf{0.4867} \\
$n=5$ & 0.2081 & 24.12 & 0.6461 & 0.7539 & 59.86 & 0.4860 \\ 
$n=6$ & 0.2080 & 24.13 & 0.6462 & 0.7538 & 59.85 & 0.4859 \\
$n=7$ & 0.2078 & 24.17 & 0.6465 & 0.7536 & 59.83 & 0.4857 \\
$n=8$ & \textbf{0.2077} & \textbf{24.18} & \textbf{0.6459} & 0.7541 & 59.85 & 0.4861 \\
$n=9$ & \textbf{0.2077} & \textbf{24.18} & 0.6457 & \textbf{0.7542} & 59.86 & 0.4862 \\ 
$n=10$ & \textbf{0.2077} & \textbf{24.18} & 0.6457 & \textbf{0.7542} & 59.86 & 0.4862 \\
\bottomrule
\end{tabular}
}
\end{table}

\section{Conclusion}
In this work, we propose an iterative hybrid reward-guided inference-time scaling method with Adaptive Frequency Steering for image super-resolution (SR) tasks, which achieves consistently superior performance compared with existing inference-time scaling methods. During each iteration of sampling, we introduce an adaptive frequency steering (AFS) strategy. The AFS mechanism achieves precise optimization of sampled particles by aligning and fusing low-frequency information of the optimal particle with high-similarity information in the particle pool. Simultaneously, the iterative hybrid reward strategy refines the results through multiple rounds of guidance. Experiments show that our method significantly outperforms existing inference-time scaling techniques in balancing perceptual-structural metrics and generates super-resolution images with both high perceptual quality and high structural fidelity. This addresses the core challenge of balancing perceptual quality and structural fidelity in image SR tasks using diffusion models. We expect our method to be generalized to other image generation tasks in the future.
{
    \small
    \bibliographystyle{ieeenat_fullname}
    \bibliography{main}

@String(AAAI = {AAAI})

@article{saharia2022image,
  title={Image super-resolution via iterative refinement},
  author={Saharia, Chitwan and Ho, Jonathan and Chan, William and Salimans, Tim and Fleet, David J and Norouzi, Mohammad},
  journal={IEEE transactions on pattern analysis and machine intelligence},
  volume={45},
  number={4},
  pages={4713--4726},
  year={2022},
  publisher={IEEE}
}

@article{li2022srdiff,
  title={Srdiff: Single image super-resolution with diffusion probabilistic models},
  author={Li, Haoying and Yang, Yifan and Chang, Meng and Chen, Shiqi and Feng, Huajun and Xu, Zhihai and Li, Qi and Chen, Yueting},
  journal={Neurocomputing},
  volume={479},
  pages={47--59},
  year={2022},
  publisher={Elsevier}
}

@inproceedings{rombach2022high,
  title={High-resolution image synthesis with latent diffusion models},
  author={Rombach, Robin and Blattmann, Andreas and Lorenz, Dominik and Esser, Patrick and Ommer, Bj{\"o}rn},
  booktitle={Proceedings of the IEEE/CVF conference on computer vision and pattern recognition},
  pages={10684--10695},
  year={2022}
}

@article{yue2023resshift,
  title={Resshift: Efficient diffusion model for image super-resolution by residual shifting},
  author={Yue, Zongsheng and Wang, Jianyi and Loy, Chen Change},
  journal={Advances in Neural Information Processing Systems},
  volume={36},
  pages={13294--13307},
  year={2023}
}

@inproceedings{lin2024diffbir,
  title={Diffbir: Toward blind image restoration with generative diffusion prior},
  author={Lin, Xinqi and He, Jingwen and Chen, Ziyan and Lyu, Zhaoyang and Dai, Bo and Yu, Fanghua and Qiao, Yu and Ouyang, Wanli and Dong, Chao},
  booktitle={European conference on computer vision},
  pages={430--448},
  year={2024},
  organization={Springer}
}

@inproceedings{wu2024seesr,
  title={Seesr: Towards semantics-aware real-world image super-resolution},
  author={Wu, Rongyuan and Yang, Tao and Sun, Lingchen and Zhang, Zhengqiang and Li, Shuai and Zhang, Lei},
  booktitle={Proceedings of the IEEE/CVF conference on computer vision and pattern recognition},
  pages={25456--25467},
  year={2024}
}

@article{wu2024one,
  title={One-step effective diffusion network for real-world image super-resolution},
  author={Wu, Rongyuan and Sun, Lingchen and Ma, Zhiyuan and Zhang, Lei},
  journal={Advances in Neural Information Processing Systems},
  volume={37},
  pages={92529--92553},
  year={2024}
}

@inproceedings{yang2024pixel,
  title={Pixel-aware stable diffusion for realistic image super-resolution and personalized stylization},
  author={Yang, Tao and Wu, Rongyuan and Ren, Peiran and Xie, Xuansong and Zhang, Lei},
  booktitle={European conference on computer vision},
  pages={74--91},
  year={2024},
  organization={Springer}
}

@article{uehara2025inference,
  title={Inference-time alignment in diffusion models with reward-guided generation: Tutorial and review},
  author={Uehara, Masatoshi and Zhao, Yulai and Wang, Chenyu and Li, Xiner and Regev, Aviv and Levine, Sergey and Biancalani, Tommaso},
  journal={arXiv preprint arXiv:2501.09685},
  year={2025}
}

@article{ma2025inference,
  title={Inference-time scaling for diffusion models beyond scaling denoising steps},
  author={Ma, Nanye and Tong, Shangyuan and Jia, Haolin and Hu, Hexiang and Su, Yu-Chuan and Zhang, Mingda and Yang, Xuan and Li, Yandong and Jaakkola, Tommi and Jia, Xuhui and others},
  journal={arXiv preprint arXiv:2501.09732},
  year={2025}
}

@article{singhal2025general,
  title={A general framework for inference-time scaling and steering of diffusion models},
  author={Singhal, Raghav and Horvitz, Zachary and Teehan, Ryan and Ren, Mengye and Yu, Zhou and McKeown, Kathleen and Ranganath, Rajesh},
  journal={arXiv preprint arXiv:2501.06848},
  year={2025}
}

@article{li2025dynamic,
  title={Dynamic Search for Inference-Time Alignment in Diffusion Models},
  author={Li, Xiner and Uehara, Masatoshi and Su, Xingyu and Scalia, Gabriele and Biancalani, Tommaso and Regev, Aviv and Levine, Sergey and Ji, Shuiwang},
  journal={arXiv preprint arXiv:2503.02039},
  year={2025}
}

@article{guo2025training,
  title={Training-free guidance beyond differentiability: Scalable path steering with tree search in diffusion and flow models},
  author={Guo, Yingqing and Yang, Yukang and Yuan, Hui and Wang, Mengdi},
  journal={arXiv preprint arXiv:2502.11420},
  year={2025}
}

@article{oshima2025inference,
  title={Inference-time text-to-video alignment with diffusion latent beam search},
  author={Oshima, Yuta and Suzuki, Masahiro and Matsuo, Yutaka and Furuta, Hiroki},
  journal={arXiv preprint arXiv:2501.19252},
  year={2025}
}

@article{zhang2025inference,
  title={Inference-time scaling of diffusion models through classical search},
  author={Zhang, Xiangcheng and Lin, Haowei and Ye, Haotian and Zou, James and Ma, Jianzhu and Liang, Yitao and Du, Yilun},
  journal={arXiv preprint arXiv:2505.23614},
  year={2025}
}

@article{dang2025inference,
  title={Inference-time scaling of diffusion language models with particle gibbs sampling},
  author={Dang, Meihua and Han, Jiaqi and Xu, Minkai and Xu, Kai and Srivastava, Akash and Ermon, Stefano},
  journal={arXiv preprint arXiv:2507.08390},
  year={2025}
}

@article{mao2025ctrl,
  title={Ctrl-Z Sampling: Diffusion Sampling with Controlled Random Zigzag Explorations},
  author={Mao, Shunqi and Guo, Wei and Zhang, Chaoyi and Long, Jieting and Xie, Ke and Cai, Weidong},
  journal={arXiv preprint arXiv:2506.20294},
  year={2025}
}

@article{jajal2025inference,
  title={Inference-Time Alignment of Diffusion Models with Evolutionary Algorithms},
  author={Jajal, Purvish and Eliopoulos, Nick John and Chou, Benjamin Shiue-Hal and Thiruvathukal, George K and Davis, James C and Lu, Yung-Hsiang},
  journal={arXiv preprint arXiv:2506.00299},
  year={2025}
}

@article{hu2025kernel,
  title={Kernel density steering: Inference-time scaling via mode seeking for image restoration},
  author={Hu, Yuyang and Mei, Kangfu and Sahraee-Ardakan, Mojtaba and Kamilov, Ulugbek S and Milanfar, Peyman and Delbracio, Mauricio},
  journal={arXiv preprint arXiv:2507.05604},
  year={2025}
}

@article{farahbakhsh2025inference,
  title={Inference-Time Search using Side Information for Diffusion-based Image Reconstruction},
  author={Farahbakhsh, Mahdi and Kunde, Vishnu Teja and Kalathil, Dileep and Narayanan, Krishna and Chamberland, Jean-Francois},
  journal={arXiv preprint arXiv:2510.03352},
  year={2025}
}

@article{su2025navigating,
  title={Navigating the Exploration-Exploitation Tradeoff in Inference-Time Scaling of Diffusion Models},
  author={Su, Xun and Huang, Jianming and Yusen, Yang and Fang, Zhongxi and Kasai, Hiroyuki},
  journal={arXiv preprint arXiv:2508.12361},
  year={2025}
}

@ARTICLE{1000236,
  author={Comaniciu, D. and Meer, P.},
  journal={IEEE Transactions on Pattern Analysis and Machine Intelligence}, 
  title={Mean shift: a robust approach toward feature space analysis}, 
  year={2002},
  volume={24},
  number={5},
  pages={603-619},
  doi={10.1109/34.1000236}}

@incollection{davis2011remarks,
  title={Remarks on some nonparametric estimates of a density function},
  author={Davis, Richard A and Lii, Keh-Shin and Politis, Dimitris N},
  booktitle={Selected Works of Murray Rosenblatt},
  pages={95--100},
  year={2011},
  publisher={Springer}
}

@article{parzen1962estimation,
  title={On estimation of a probability density function and mode},
  author={Parzen, Emanuel},
  journal={The annals of mathematical statistics},
  volume={33},
  number={3},
  pages={1065--1076},
  year={1962},
  publisher={JSTOR}
}

@inproceedings{blau2018perception,
  title={The perception-distortion tradeoff},
  author={Blau, Yochai and Michaeli, Tomer},
  booktitle={Proceedings of the IEEE conference on computer vision and pattern recognition},
  pages={6228--6237},
  year={2018}
}

@inproceedings{ledig2017photo,
  title={Photo-realistic single image super-resolution using a generative adversarial network},
  author={Ledig, Christian and Theis, Lucas and Husz{\'a}r, Ferenc and Caballero, Jose and Cunningham, Andrew and Acosta, Alejandro and Aitken, Andrew and Tejani, Alykhan and Totz, Johannes and Wang, Zehan and others},
  booktitle={Proceedings of the IEEE conference on computer vision and pattern recognition},
  pages={4681--4690},
  year={2017}
}

@article{valentini2017best,
  title={The best-of-n problem in robot swarms: Formalization, state of the art, and novel perspectives},
  author={Valentini, Gabriele and Ferrante, Eliseo and Dorigo, Marco},
  journal={Frontiers in Robotics and AI},
  volume={4},
  pages={9},
  year={2017},
  publisher={Frontiers Media SA}
}

@article{song2020denoising,
  title={Denoising diffusion implicit models},
  author={Song, Jiaming and Meng, Chenlin and Ermon, Stefano},
  journal={arXiv preprint arXiv:2010.02502},
  year={2020}
}

@article{kingma2021variational,
  title={Variational diffusion models},
  author={Kingma, Diederik and Salimans, Tim and Poole, Ben and Ho, Jonathan},
  journal={Advances in neural information processing systems},
  volume={34},
  pages={21696--21707},
  year={2021}
}

@article{song2019generative,
  title={Generative modeling by estimating gradients of the data distribution},
  author={Song, Yang and Ermon, Stefano},
  journal={Advances in neural information processing systems},
  volume={32},
  year={2019}
}

@inproceedings{huang2017arbitrary,
  title={Arbitrary style transfer in real-time with adaptive instance normalization},
  author={Huang, Xun and Belongie, Serge},
  booktitle={Proceedings of the IEEE international conference on computer vision},
  pages={1501--1510},
  year={2017}
}

@inproceedings{steinbiss1994improvements,
  title={Improvements in beam search.},
  author={Steinbiss, Volker and Tran, Bach-Hiep and Ney, Hermann},
  booktitle={ICSLP},
  volume={94},
  pages={2143--2146},
  year={1994}
}

@inproceedings{agustsson2017ntire,
  title={Ntire 2017 challenge on single image super-resolution: Dataset and study},
  author={Agustsson, Eirikur and Timofte, Radu},
  booktitle={Proceedings of the IEEE conference on computer vision and pattern recognition workshops},
  pages={126--135},
  year={2017}
}

@inproceedings{cai2019toward,
  title={Toward real-world single image super-resolution: A new benchmark and a new model},
  author={Cai, Jianrui and Zeng, Hui and Yong, Hongwei and Cao, Zisheng and Zhang, Lei},
  booktitle={Proceedings of the IEEE/CVF international conference on computer vision},
  pages={3086--3095},
  year={2019}
}

@inproceedings{wei2020component,
  title={Component divide-and-conquer for real-world image super-resolution},
  author={Wei, Pengxu and Xie, Ziwei and Lu, Hannan and Zhan, Zongyuan and Ye, Qixiang and Zuo, Wangmeng and Lin, Liang},
  booktitle={European conference on computer vision},
  pages={101--117},
  year={2020},
  organization={Springer}
}

@inproceedings{deng2009imagenet,
  title={Imagenet: A large-scale hierarchical image database},
  author={Deng, Jia and Dong, Wei and Socher, Richard and Li, Li-Jia and Li, Kai and Fei-Fei, Li},
  booktitle={2009 IEEE conference on computer vision and pattern recognition},
  pages={248--255},
  year={2009},
  organization={Ieee}
}

@inproceedings{zhang2018unreasonable,
  title={The unreasonable effectiveness of deep features as a perceptual metric},
  author={Zhang, Richard and Isola, Phillip and Efros, Alexei A and Shechtman, Eli and Wang, Oliver},
  booktitle={Proceedings of the IEEE conference on computer vision and pattern recognition},
  pages={586--595},
  year={2018}
}

@inproceedings{ke2021musiq,
  title={Musiq: Multi-scale image quality transformer},
  author={Ke, Junjie and Wang, Qifei and Wang, Yilin and Milanfar, Peyman and Yang, Feng},
  booktitle={Proceedings of the IEEE/CVF international conference on computer vision},
  pages={5148--5157},
  year={2021}
}

@inproceedings{wang2023exploring,
  title={Exploring clip for assessing the look and feel of images},
  author={Wang, Jianyi and Chan, Kelvin CK and Loy, Chen Change},
  booktitle={Proceedings of the AAAI conference on artificial intelligence},
  volume={37},
  number={2},
  pages={2555--2563},
  year={2023}
}

@inproceedings{yang2022maniqa,
  title={Maniqa: Multi-dimension attention network for no-reference image quality assessment},
  author={Yang, Sidi and Wu, Tianhe and Shi, Shuwei and Lao, Shanshan and Gong, Yuan and Cao, Mingdeng and Wang, Jiahao and Yang, Yujiu},
  booktitle={Proceedings of the IEEE/CVF conference on computer vision and pattern recognition},
  pages={1191--1200},
  year={2022}
}

@article{wu2025one,
  title={One-Step Diffusion-based Real-World Image Super-Resolution with Visual Perception Distillation},
  author={Wu, Xue and Xin, Jingwei and Tu, Zhijun and Hu, Jie and Li, Jie and Wang, Nannan and Gao, Xinbo},
  journal={arXiv preprint arXiv:2506.02605},
  year={2025}
}

@inproceedings{liao2022deepwsd,
  title={DeepWSD: Projecting degradations in perceptual space to wasserstein distance in deep feature space},
  author={Liao, Xingran and Chen, Baoliang and Zhu, Hanwei and Wang, Shiqi and Zhou, Mingliang and Kwong, Sam},
  booktitle={Proceedings of the 30th ACM International Conference on Multimedia},
  pages={970--978},
  year={2022}
}

@article{xia2015learning,
  title={Learning similarity with cosine similarity ensemble},
  author={Xia, Peipei and Zhang, Li and Li, Fanzhang},
  journal={Information sciences},
  volume={307},
  pages={39--52},
  year={2015},
  publisher={Elsevier}
}

@article{yang2010image,
  title={Image super-resolution via sparse representation},
  author={Yang, Jianchao and Wright, John and Huang, Thomas S and Ma, Yi},
  journal={IEEE transactions on image processing},
  volume={19},
  number={11},
  pages={2861--2873},
  year={2010},
  publisher={IEEE}
}

@inproceedings{glasner2009super,
  title={Super-resolution from a single image},
  author={Glasner, Daniel and Bagon, Shai and Irani, Michal},
  booktitle={2009 IEEE 12th international conference on computer vision},
  pages={349--356},
  year={2009},
  organization={IEEE}
}

@article{dong2015image,
  title={Image super-resolution using deep convolutional networks},
  author={Dong, Chao and Loy, Chen Change and He, Kaiming and Tang, Xiaoou},
  journal={IEEE transactions on pattern analysis and machine intelligence},
  volume={38},
  number={2},
  pages={295--307},
  year={2015},
  publisher={IEEE}
}

@article{liang2024scaling,
  title={Scaling laws for diffusion transformers},
  author={Liang, Zhengyang and He, Hao and Yang, Ceyuan and Dai, Bo},
  journal={arXiv preprint arXiv:2410.08184},
  year={2024}
}

@article{fang2024vivid,
  title={Vivid: Video virtual try-on using diffusion models},
  author={Fang, Zixun and Zhai, Wei and Su, Aimin and Song, Hongliang and Zhu, Kai and Wang, Mao and Chen, Yu and Liu, Zhiheng and Cao, Yang and Zha, Zheng-Jun},
  journal={arXiv preprint arXiv:2405.11794},
  year={2024}
}
}
\clearpage
\setcounter{page}{1}
\maketitlesupplementary

\section{Theoretical Analysis}
In this section, we present a rigorous derivation of Eq.~\ref{4} and Eq.~\ref{5} from the main text, providing detailed mathematical steps that were omitted for brevity. We also visualize the spectral evolution of the intermediate latent variables across the reverse diffusion process, as shown in Fig.~\ref{fig:spectral}. These curves reveal how frequency components are progressively reconstructed from noise to image structure. Leveraging the spectral decay characteristics of Gaussian perturbations, we further establish the theoretical foundation for the proposed Adaptive Frequency Steering (AFS) strategy. Together, these analyses offer a comprehensive understanding of the frequency behavior underlying diffusion inference.
\subsection{Justification for Frequency Split}
Based on the spectral decay properties of Gaussian noise, we theoretically justify the rationality of the proposed Adaptive Frequency Steering (AFS) strategy. As defined in Eq.~\ref{eq:posterior} of the main text, the reverse denoising process $\mathbf{p_{\theta}(\mathbf{x}{t-1} \mid \mathbf{x}{t})}$ is modeled as a Gaussian distribution, which is a common formulation in diffusion-based generative models. To analyze the behavior of this process in the frequency domain, we examine the Fourier transform of its probability density function (PDF). This transformation allows us to investigate how uncertainty propagates across different frequency components, which is essential for understanding the selective attenuation or preservation mechanisms that AFS leverages. In probability theory, the Fourier transform of a PDF is referred to as the Characteristic Function, serving as a compact representation that encodes all distributional moments and provides a convenient tool for frequency-domain reasoning. For a $d$-dimensional multivariate Gaussian distribution $X \sim \mathcal{N}(\mu, \Sigma)$, the probability density function (PDF) is given by:
\begin{equation}
f(x) = \frac{1}{\sqrt{(2\pi)^d |\Sigma|}}
\exp\!\left(
    -\frac{1}{2} (x - \mu)^{\top} \Sigma^{-1} (x - \mu)
\right).
\label{ap:1}
\end{equation}
The Fourier transform is defined as 
$\hat{f}(\omega) = \mathbb{E}[e^{-i\omega^{\top} x}]$. 
Applying this to the Gaussian distribution yields:
\begin{equation}
\begin{aligned}
\hat{f}_{x_{t-1}}(\omega)
&= \int_{\mathbb{R}^d} p_{\theta}(x)\, e^{-i\omega^{\top} x}\, dx \\[4pt]
&= \exp\!\left(
    -i\omega^{\top}\mu_{\theta}
    - \frac{1}{2}\omega^{\top}\Sigma\,\omega
\right).
\end{aligned}
\label{ap:2}
\end{equation}
In our diffusion setting, the covariance matrix represents isotropic Gaussian noise, where $\Sigma = \sigma_t^2 I$. Substituting this into the equation above yields Eq.~\ref{4} from the main text:
\begin{equation}
\hat{f}_{x_{t-1}}(\omega)
= 
\exp\!\left(
    -i \omega^{\top} \mu_{\theta}
    - \frac{1}{2} \sigma_t^2 \|\omega\|^2
\right).
\label{ap:3}
\end{equation}
To analyze the energy distribution across frequencies, we focus on the Magnitude Spectrum. Utilizing the property of complex moduli 
$|e^{i\theta}| = 1$, the magnitude of the phase term 
$\exp(-i\omega^{\top}\mu_{\theta})$ is unity. Consequently, the amplitude spectrum is determined by the real exponential component:
\begin{equation}
\begin{aligned}
| \hat{f}(\omega) |
&= \left| \exp(-i\omega^{\top}\mu_{\theta}) \right|
   \cdot 
   \left| 
       \exp\!\left( -\frac{1}{2}\sigma_t^2 \|\omega\|^2 \right)
   \right| \\[4pt]
&= 1 \cdot 
   \exp\!\left( -\frac{1}{2}\sigma_t^2 \|\omega\|^2 \right) \\[4pt]
&= \exp\!\left( -\frac{1}{2}\sigma_t^2 \|\omega\|^2 \right).
\label{ap:4}
\end{aligned}
\end{equation}
This confirms Eq.~\ref{5} in the main text.
\subsection{Frequency Evolution in Inference Process}
\begin{figure}[H]
    \centering 
    \includegraphics[width=1.0\linewidth]{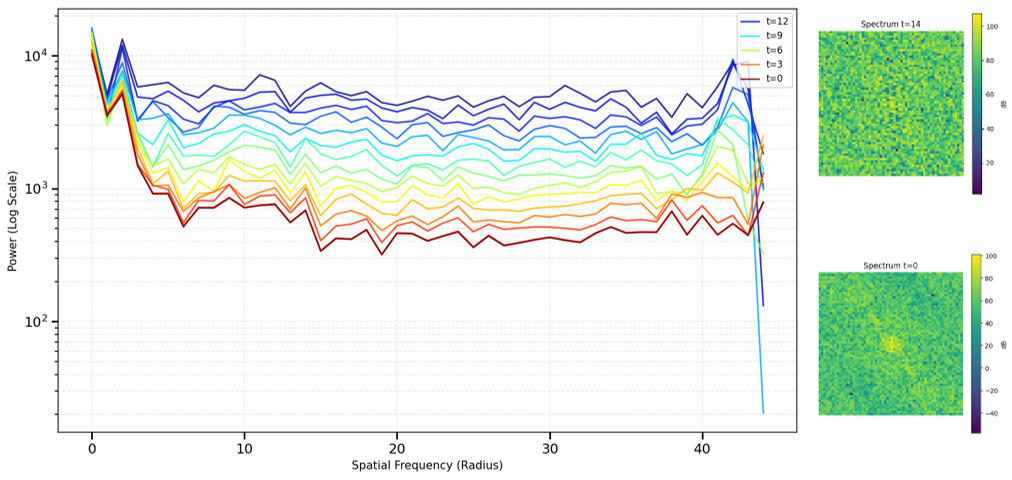} 
    \caption{Evolution of the 1D Power Spectral Density (PSD) during the reverse diffusion process. } % 图片标题
    \label{fig:spectral} % 标签
\end{figure}

We compute the Radially Averaged Power Spectral Density (RAPSD) for the sampled trajectories. Specifically, we sample Gaussian noise $x_T$ and perform the reverse denoising steps using the pre-trained ResShift backbone. For each timestep $t$, we apply a 2D Discrete Fourier Transform (DFT) to $x_t$, compute the magnitude spectrum, and then perform azimuthal integration to obtain the 1D PSD profile.
Fig.~\ref{fig:spectral} illustrates the evolution of the 1D PSD curves from $t=14$ (initial noise state) to $t=0$ (final generated image). The X-axis represents spatial frequency (radius), and the Y-axis represents the power magnitude in a logarithmic scale.
The empirical analysis of spectral evolution reveals a clear temporal disparity in feature reconstruction. Low-frequency components rapidly converge to the target distribution, establishing the global structure during the early stages of inference. In contrast, high-frequency components exhibit prolonged energy fluctuations, with fine-grained details recovering only in the final steps. These observations provide a solid physical interpretation for the proposed Adaptive Frequency Steering (AFS) strategy: the early-stage stability of low-frequency information justifies its use as Pseudo-GT for structural guidance, whereas the delayed recovery 
of high-frequency textures necessitates an iterative refinement mechanism. Together, these properties enable AFS to effectively balance structural fidelity and perceptual quality.
\section{Implementation Details}
In this section, we present supplementary derivations and clarifications that support the formulations introduced in the main paper. We first provide additional details on the frequency decoupling mechanism within the AFS framework. In addition, we supplement and compare different reward scheduling strategies in the particle optimization process.
\subsection{Spatial Frequency Decomposition}
In Sec.~\ref{sec:formatting} of the main text, we formulated the Adaptive Frequency Steering (AFS) mechanism using Discrete Fourier Transform (DFT) and spectral masking (Eq.~\ref{7}, Eq.~\ref{8}) to provide a clear theoretical motivation for frequency separation. However, in practical deployment, direct ideal low-pass filtering in the frequency domain often introduces undesirable Gibbs phenomena in the spatial domain. Therefore, we adopt Spatial Gaussian Filtering as an efficient and robust approximation of frequency-domain low-pass filtering in our implementation. By the convolution theorem, spatial convolution with a Gaussian kernel is equivalent to element-wise multiplication with a Gaussian function in the frequency domain. This approach not only achieves a "soft" frequency split that avoids ringing artifacts but also effectively preserves the spatial continuity of features.
Specifically, given the latent variable $x_t$ at timestep $t$, we extract the low-frequency component $x_{\text{low}}$ and the high-frequency component $x_{\text{high}}$ as follows:
\begin{equation}
x_{\text{low}} = \mathcal{G}(x_t;\, k, \sigma),
\qquad
x_{\text{high}} = x_t - x_{\text{low}},
\label{ap:5}
\end{equation}
where $\mathcal{G}(\cdot)$ denotes a Gaussian blur operator. Based on empirical tuning, we set the kernel 
size to $k = 9$ and the standard deviation to $\sigma = 1.0$. This configuration effectively suppresses 
high-frequency noise while preserving essential structural information, thereby providing a stable foundation 
for subsequent AdaIN-based alignment.

\subsection{Hybrid Reward Scheduling}
This section details the reward-scheduling strategy employed during the iterative refinement process ($n > 1$). In iterative refinement, each subsequent iteration builds upon the predictions of the previous one, making the design of the reward schedule crucial for achieving a stable and progressively improving optimization trajectory. We conducted experiments using the ResShift baseline on the ImageNet~\cite{deng2009imagenet} dataset, with the ResShift~\cite{yue2023resshift} inference time step set to 15. This setting ensures a consistent evaluation protocol across different reward configurations, allowing us to isolate the influence of reward scheduling itself.
Once a pseudo-GT is obtained from the initial iteration, the primary objective shifts to balancing the preservation of structural fidelity with the continuous enhancement of perceptual quality. To identify the optimal integration strategy for these conflicting objectives, we investigated four distinct reward scheduling configurations.
Let $\mathcal{R}_{\text{C}}$ denote the perceptual reward (CLIPIQA~\cite{wang2023exploring}) and $\mathcal{R}_{\text{L}}$ denote the structural penalty (LPIPS~\cite{zhang2018unreasonable}). We evaluated the following settings:
\begin{enumerate}
    \item \textbf{LPIPS-Only:} The guidance relies solely on minimizing the perceptual distance to the pseudo-GT throughout the entire reverse process ($r = -\mathcal{R}_{\text{L}}$).
    \item \textbf{Constant Hybrid:} A fixed weighted combination of perception and structure is applied at every timestep ($r = \mathcal{R}_{\text{C}} - \mathcal{R}_{\text{L}}$).
    \item \textbf{Linear Scheduling:} The weight of the perceptual term linearly increases while the structural term decreases as $t \to 0$, promoting structure early and texture late ($r = \alpha_t \mathcal{R}_{\text{C}} - (1-\alpha_t) \mathcal{R}_{\text{L}}$).
    \item \textbf{Segmented Scheduling:} A staged approach where the inference process is divided into three phases: structural alignment, transitional fusion, and perceptual refinement.
\end{enumerate}
As presented in Tab.~\ref{tab:reward-schedule}, the experimental result reveals that the LPIPS-only strategy exhibits a clear phenomenon of reward hacking. While it achieves the lowest LPIPS score (0.1976), this metric over-optimization comes at a severe cost to other dimensions. It yields the lowest PSNR (23.24 dB) and SSIM (0.6435), indicating that enforcing strict feature-space constraints throughout the entire inference process overly restricts the generative diversity, preventing the model from hallucinating necessary high-frequency details (evidenced by the lowest CLIPIQA score of 0.7106). The CLIPIQA-LPIPS and Linear strategies, which incorporate perceptual guidance earlier or continuously, tend to bias the generation towards visual pleasantness at the expense of fidelity. Although they achieve superior CLIPIQA scores (0.7539 for Constant), their structural metrics (PSNR/SSIM) are notably suppressed compared to segmented method. In contrast, the segmented strategy maintains competitive perceptual performance without succumbing to the artifacts associated with reward hacking or structural degradation. This confirms that decoupling structural and perceptual guidance into separate temporal windows is essential for generating photorealistic images that remain faithful to the input.
\begin{table}[H]
\centering
\caption{Performance of different reward scheduling strategies.}
\label{tab:reward-schedule}
\resizebox{\columnwidth}{!}{
\setlength{\tabcolsep}{4pt}
\renewcommand{\arraystretch}{1.05}
\footnotesize
\begin{tabular}{ccccccc}
\toprule
Reward Schedule & PSNR & SSIM & LPIPS($\downarrow$) & CLIPIQA & MUSIQ & MANIQA \\
\midrule
LPIPS           & 23.24 & 0.6435 & \textbf{0.1976} & 0.7106 & 61.44	& 0.5190\\
CLIPIQA-LPIPS   & 23.73 & 0.6475 & 0.2043 & \textbf{0.7539} & 60.68	& 0.5195 \\
Linear          & 23.73 & 0.6466 & 0.2042 & 0.7368 & 60.26 & \textbf{0.5313}\\
Segmented       & \textbf{24.13} & \textbf{0.6485} & 0.2089 & 0.7312 & \textbf{61.02} & 0.5075\\
\bottomrule
\end{tabular}
}
\end{table}

\section{Additional Ablation Studies}
In this section, we provide a comprehensive analysis of the core design choices within our IAFS framework. We first investigate the impact of different perceptual metrics when serving as the guidance reward during the initial iteration. Subsequently, we conduct a fine-grained search to identify the optimal temporal thresholds for our segmented scheduling strategy.
\subsection{Ablation on Perception Reward}
The quality of the pseudo-ground truth (pseudo-GT) generated in the first iteration ($n=1$) is pivotal for the stability of subsequent refinements. Since this initial step relies solely on a non-reference perceptual reward, the choice of the metric directly influences the initial structural foundation. To determine the most effective guidance signal, we evaluated three state-of-the-art non-reference image quality assessment (IQA) metrics as the optimization objective: CLIPIQA, MUSIQ~\cite{ke2021musiq}, and MANIQA~\cite{yang2022maniqa}.
The quantitative results are summarized in Tab.~\ref{tab:perception}, we can observe that MUSIQ and MANIQA achieve the highest scores on their respective metrics (67.76 for MUSIQ and 0.6049 for MANIQA). However, this comes at a significant cost to structural fidelity and general perceptual quality. Specifically, using MANIQA results in a PSNR drop of nearly 1 dB (23.10 dB vs. 24.05 dB) and a degradation in LPIPS compared to CLIPIQA. These findings indicate that while specialized IQA metrics like MUSIQ and MANIQA are effective for evaluation, CLIPIQA offers better generalization and alignment with both perception and structural consistency when used as an optimization objective. Consequently, we adopt CLIPIQA as the default perceptual reward to ensure a high-quality starting point for the iterative refinement process.
\begin{table}[H]
\centering
\caption{Performance of different perceptual rewards in the first iteration.}
\label{tab:perception}
\resizebox{\columnwidth}{!}{
\setlength{\tabcolsep}{4pt}
\renewcommand{\arraystretch}{1.05}
\footnotesize
\begin{tabular}{ccccccc}
\toprule
Reward & LPIPS($\downarrow$) & PSNR & SSIM & CLIPIQA & MUSIQ & MANIQA \\
\midrule
CLIPIQA  & \textbf{0.2105} & \textbf{24.05} & 0.6450 & \textbf{0.7504} & 59.71 & 0.4837 \\
MUSIQ    & 0.2134 & 23.29 & \textbf{0.6464} & 0.7158 & \textbf{67.76} & 0.5385 \\
MANIQA   & 0.2124 & 23.10 & 0.6418 & 0.7035 & 60.97 & \textbf{0.6049} \\
\bottomrule
\end{tabular}
}
\end{table}

\begin{table*}[t!]
\caption{Comparison of time and memory under different methods in the first iteration and the next iteration.}
\label{tab:time}
\centering
\setlength{\tabcolsep}{10pt}
\renewcommand{\arraystretch}{1.2}
\resizebox{0.9\linewidth}{!}{%
\begin{tabular}{c|cc|cc|cc|cc|cc}
\toprule
\multirow{2}{*}{Method} &
\multicolumn{2}{c|}{BON} &
\multicolumn{2}{c|}{BS}  &
\multicolumn{2}{c|}{FK}  &
\multicolumn{2}{c|}{KDS} &
\multicolumn{2}{c}{Ours} \\
& n=1 & n=2 & n=1 & n=2 & n=1 & n=2 & n=1 & n=2 & n=1 & n=2 \\
\midrule
Time (s) &
5.002s & 5.020s &
47.17s & 55.58s &
49.49s & 53.11s &
131.7s & 131.7s &
50.10s & 58.62s \\
Memory (Gb) &
6.00Gb & 6.00Gb &
6.00Gb & 6.00Gb &
6.00Gb & 6.00Gb &
3.87Gb & 3.87Gb &
6.00Gb & 6.00Gb \\
\bottomrule
\end{tabular}
}
\end{table*}

\subsection{Ablation on Transition Thresholds}
\begin{figure}[H]
    \centering 
    \includegraphics[width=0.8\linewidth]{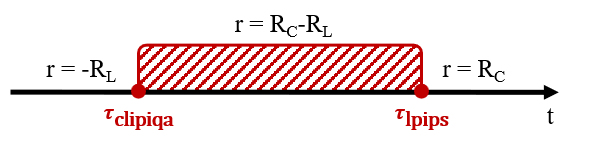} 
    \caption{The segmented reward scheduling strategy with $\tau_{\text{clipiqa}}$ and $\tau_{\text{lpips}}$. } % 图片标题
    \label{fig:reward} % 标签
\end{figure}

To further optimize the Segmented Scheduling strategy, we conducted an ablation study on the specific time thresholds that define the three phases. We specifically analyze two key hyperparameters: the start timestep for introducing CLIPIQA ($\tau_{\text{clipiqa}}$) and the end timestep for removing LPIPS ($\tau_{\text{lpips}}$). The segmented reward scheduling strategy is shown in Fig.~\ref{fig:reward}.
We evaluated multiple combinations of $[\tau_{clipiqa}, \tau_{lpips}]$ on a randomly sampled subset of 1,000 images from the ImageNet validation set using the ResShift backbone ($T=15$). The quantitative results are comprehensively summarized in Tab.~\ref{tab:clip-lpips}. Comparing the sub-tables where $\tau_{\text{clipiqa}}$ ranges from 0 to 9, we observe a distinct trade-off. When $\tau_{\text{clipiqa}}$ is small (e.g., $\tau_{\text{clipiqa}} \in [0, 2]$), meaning perceptual guidance is introduced very late in the denoising process, the model achieves high structural fidelity (SSIM $\approx 0.652$) but suboptimal perceptual quality. This indicates that intervening only in the final few steps is insufficient to hallucinate vivid high-frequency details. Conversely, setting $\tau_{\text{clipiqa}}$ too high (e.g., $\tau_{\text{clipiqa}} \ge 8$) introduces perceptual gradients while the image is still heavily corrupted by noise. 
For a fixed $\tau_{\text{clipiqa}}$, varying $\tau_{\text{lpips}}$ controls the duration of the mixed guidance phase. The results show that retaining the LPIPS constraint for too long prevents the model from generating sharp textures, as evidenced by lower MANIQA and MUSIQ scores. However, removing it too early causes the generation to drift away from the pseudo-GT, harming fidelity. By analyzing the Pareto frontier of structural fidelity (PSNR/SSIM) and perceptual quality (LPIPS/CLIPIQA), we identify the configuration $[\tau_{\text{clipiqa}}=4, \tau_{\text{lpips}}=7]$ as the optimal operating point.
Under this setting, the model achieves a PSNR of 24.13 dB and an SSIM of 0.6485, which are among the highest across all combinations. Simultaneously, it maintains excellent perceptual performance with an LPIPS of 0.2089 and a competitive CLIPIQA score. This specific schedule effectively utilizes the mid-inference stage ($t \in [4, 7]$) to transition from structural correction to texture refinement, ensuring a robust balance between faithfulness and realism. Consequently, we adopt $[\tau_{\text{clipiqa}}=4, \tau_{\text{lpips}}=7]$ as the default setting for all subsequent experiments.
\begin{table*}[t!]
\caption{Quantitative Comparison of Different Combinations of $[\tau_{\text{clipiqa}}, \tau_{\text{lpips}}]$. The best and second best results are marked in \textcolor{red}{red} and \textcolor{blue}{blue}.}
\label{tab:clip-lpips}
\centering
% 设置行距和列间距
\renewcommand{\arraystretch}{1.1} 
\setlength{\tabcolsep}{2pt}

% 使用 resizebox 强制表格适应页面宽度
\resizebox{0.9\linewidth}{!}{%
\begin{tabular}{c|cccccc|c|cccccc}
\hline
% =========================================================
% Block 1: clip=0 (Left) vs clip=1 (Right)
% =========================================================
\multirow{2}{*}{$\tau_{\text{lpips}}$} & \multicolumn{6}{c|}{$\tau_{\text{clipiqa}}=0$} & \multirow{2}{*}{$\tau_{\text{lpips}}$} & \multicolumn{6}{c}{$\tau_{\text{clipiqa}}=1$} \\
 & PSNR & SSIM & LPIPS($\downarrow$) & CLIPIQA & MUSIQ & MANIQA &  & PSNR & SSIM & LPIPS($\downarrow$) & CLIPIQA & MUSIQ & MANIQA \\ \hline

1 & 24.05 & 0.6493 & 0.2056 & \textcolor{red}{0.7679} & \textcolor{red}{64.49} & 0.5004 & 
1 & 24.15 & 0.6523 & 0.2128 & 0.7699 & 59.16 & 0.4828 \\

2 & 24.05 & \textcolor{red}{0.6498} & 0.2051 & \textcolor{blue}{0.7602} & 60.44 & 0.4983 & 
2 & 24.15 & 0.6523 & 0.2128 & 0.7699 & 59.16 & 0.4828 \\

3 & 24.05 & 0.6495 & 0.2047 & 0.7388 & 60.65 & \textcolor{blue}{0.5012} & 
3 & 24.15 & 0.6522 & 0.2123 & 0.7699 & 58.99 & 0.4791 \\

4 & \textcolor{red}{24.09} & \textcolor{blue}{0.6497} & 0.2075 & 0.7243 & 60.65 & 0.4897 & 
4 & 24.15 & \textcolor{blue}{0.6525} & 0.2135 & 0.7690 & 58.95 & 0.4761 \\

5 & \textcolor{blue}{24.06} & 0.6492 & 0.2058 & 0.7203 & 60.58 & 0.4908 & 
5 & 24.15 & \textcolor{red}{0.6526} & 0.2132 & 0.7694 & 59.03 & 0.4777 \\

6 & 24.05 & 0.6485 & 0.2054 & 0.7265 & 60.92 & 0.4912 & 
6 & \textcolor{red}{24.18} & 0.6518 & 0.2114 & 0.7683 & 59.13 & 0.4842 \\

% 左边第7行是灰底，右边正常
\cellcolor{gray!25}7 & \cellcolor{gray!25}24.04 & \cellcolor{gray!25}0.6492 & \cellcolor{gray!25}0.2045 & \cellcolor{gray!25}0.7252 & \cellcolor{gray!25}\textcolor{blue}{61.11} & \cellcolor{gray!25}0.4944 & 
7 & \textcolor{blue}{24.16} & 0.6520 & \textcolor{blue}{0.2110} & 0.7689 & 59.26 & \textcolor{blue}{0.4849} \\

% 左边正常，右边第8行是灰底
8 & 24.05 & 0.6486 & 0.2037 & 0.7269 & 60.32 & 0.4967 & 
\cellcolor{gray!25}8 & \cellcolor{gray!25}24.15 & \cellcolor{gray!25}0.6520 & \cellcolor{gray!25}\textcolor{red}{0.2109} & \cellcolor{gray!25}0.7718 & \cellcolor{gray!25}59.33 & \cellcolor{gray!25}0.4840 \\

9 & 24.04 & 0.6477 & 0.2018 & 0.6976 & 60.56 & 0.4873 & 
9 & 24.14 & 0.6516 & 0.2113 & 0.7735 & \textcolor{blue}{59.45} & \textcolor{red}{0.4879} \\

10 & 24.02 & 0.6459 & 0.2018 & 0.7326 & 60.53 & \textcolor{blue}{0.5069} & 
10 & 24.11 & 0.6513 & 0.2118 & 0.7727 & \textcolor{red}{59.50} & 0.4837 \\

11 & 23.93 & 0.6461 & \textcolor{blue}{0.2012} & 0.7234 & 60.26 & 0.4973 & 
11 & 24.11 & 0.6512 & 0.2125 & \textcolor{blue}{0.7753} & 59.19 & 0.4829 \\

12 & 23.95 & 0.6449 & \textcolor{red}{0.1994} & 0.7182 & 59.64 & 0.4991 & 
12 & 24.11 & 0.6511 & 0.2130 & \textcolor{red}{0.7760} & 59.08 & 0.4805 \\

\hline
% =========================================================
% Block 2: clip=2 (Left) vs clip=3 (Right)
% =========================================================
\multirow{2}{*}{$\tau_{\text{lpips}}$} & \multicolumn{6}{c|}{$\tau_{\text{clipiqa}}=2$} & \multirow{2}{*}{$\tau_{\text{lpips}}$} & \multicolumn{6}{c}{$\tau_{\text{clipiqa}}=3$} \\
 & PSNR & SSIM & LPIPS($\downarrow$) & CLIPIQA & MUSIQ & MANIQA &  & PSNR & SSIM & LPIPS($\downarrow$) & CLIPIQA & MUSIQ & MANIQA \\ \hline

2 & 24.15 & 0.6522 & 0.2124 & 0.7416 & 59.18 & 0.4782 & 
2 & 24.16 & 0.6524 & 0.2141 & 0.7300 & 58.79 & 0.4756 \\

3 & 24.14 & 0.6521 & 0.2129 & 0.7417 & 59.09 & 0.4760 & 
3 & 24.16 & 0.6524 & 0.2141 & 0.7300 & 58.79 & 0.4756 \\

4 & 24.14 & \textcolor{blue}{0.6525} & 0.2130 & 0.7433 & 59.19 & 0.4740 & 
4 & \textcolor{red}{24.19} & \textcolor{blue}{0.6529} & 0.2144 & 0.7302 & 58.86 & 0.4740 \\

5 & \textcolor{red}{24.18} & \textcolor{red}{0.6526} & 0.2126 & 0.7501 & 59.34 & 0.4767 & 
5 & \textcolor{blue}{24.18} & \textcolor{red}{0.6531} & 0.2141 & 0.7276 & 59.06 & 0.4709 \\

6 & \textcolor{blue}{24.16} & 0.6519 & 0.2107 & 0.7495 & 59.45 & 0.4822 & 
6 & 24.17 & 0.6525 & 0.2122 & 0.7277 & 59.09 & 0.4770 \\

% 左边第7行灰底
\cellcolor{gray!25}7 & \cellcolor{gray!25}24.15 & \cellcolor{gray!25}0.6521 & \cellcolor{gray!25}\textcolor{red}{0.2101} & \cellcolor{gray!25}0.7535 & \cellcolor{gray!25}59.49 & \cellcolor{gray!25}\textcolor{blue}{0.4839} & 
7 & 24.16 & 0.6522 & \textcolor{red}{0.2115} & 0.7299 & \textcolor{blue}{59.14} & \textcolor{blue}{0.4803} \\

% 右边第8行灰底
8 & 24.14 & 0.6519 & \textcolor{blue}{0.2103} & 0.7559 & 59.56 & \textcolor{blue}{0.4839} & 
\cellcolor{gray!25}8 & \cellcolor{gray!25}24.13 & \cellcolor{gray!25}0.6520 & \cellcolor{gray!25}\textcolor{blue}{0.2120} & \cellcolor{gray!25}0.7311 & \cellcolor{gray!25}\textcolor{red}{59.19} & \cellcolor{gray!25}0.4802 \\

9 & 24.11 & 0.6515 & 0.2105 & 0.7609 & \textcolor{blue}{59.60} & \textcolor{red}{0.4858} & 
9 & 24.12 & 0.6515 & 0.2123 & 0.7398 & 58.99 & \textcolor{red}{0.4820} \\

10 & 24.11 & 0.6512 & 0.2111 & 0.7609 & \textcolor{red}{59.62} & 0.4819 & 
10 & 24.12 & 0.6512 & 0.2130 & 0.7384 & 59.02 & 0.4796 \\

11 & 24.11 & 0.6512 & 0.2119 & \textcolor{blue}{0.7630} & 59.41 & 0.4798 & 
11 & 24.12 & 0.6509 & 0.2135 & \textcolor{red}{0.7438} & 58.76 & 0.4779 \\

12 & 24.11 & 0.6511 & 0.2123 & \textcolor{red}{0.7639} & 59.31 & 0.4790 & 
12 & 24.13 & 0.6510 & 0.2140 & \textcolor{blue}{0.7433} & 58.57 & 0.4771 \\

\hline
% =========================================================
% Block 3: clip=4 (Left) vs clip=5 (Right)
% =========================================================
\multirow{2}{*}{$\tau_{\text{lpips}}$} & \multicolumn{6}{c|}{$\tau_{\text{clipiqa}}=4$} & \multirow{2}{*}{$\tau_{\text{lpips}}$} & \multicolumn{6}{c}{$\tau_{\text{clipiqa}}=5$} \\
 & PSNR & SSIM & LPIPS($\downarrow$) & CLIPIQA & MUSIQ & MANIQA &  & PSNR & SSIM & LPIPS($\downarrow$) & CLIPIQA & MUSIQ & MANIQA \\ \hline

% 左边从4开始，右边从5开始，所以第一行右边是空的
4 & \textcolor{red}{24.13} & \textcolor{red}{0.6493} & 0.2108 & 0.7284 & 60.27 & 0.5034 & 
 & & & & & & \\ 

5 & \textcolor{blue}{24.12} & 0.6488 & 0.2099 & 0.7273 & 60.42 & 0.5019 & 
5 & \textcolor{blue}{24.19} & \textcolor{red}{0.6542} & 0.2125 & 0.7152 & 59.06 & \textcolor{blue}{0.4694} \\

6 & \textcolor{blue}{24.12} & \textcolor{blue}{0.6490} & 0.2098 & 0.7267 & 60.82 & 0.5041 & 
6 & \textcolor{blue}{24.19} & 0.6538 & 0.2110 & 0.7151 & \textcolor{blue}{59.10} & \textcolor{red}{0.4701} \\

% 左右第7行都是灰底
\cellcolor{gray!25}7 & \cellcolor{gray!25}\textcolor{red}{24.13} & \cellcolor{gray!25}0.6485 & \cellcolor{gray!25}0.2089 & \cellcolor{gray!25}0.7312 & \cellcolor{gray!25}61.02 & \cellcolor{gray!25}0.5075 & 
\cellcolor{gray!25}7 & \cellcolor{gray!25}24.20 & \cellcolor{gray!25}\textcolor{blue}{0.6540} & \cellcolor{gray!25}\textcolor{red}{0.2095} & \cellcolor{gray!25}0.7198 & \cellcolor{gray!25}\textcolor{red}{59.16} & \cellcolor{gray!25}0.4668 \\

8 & 24.10 & 0.6481 & 0.2079 & 0.7321 & \textcolor{red}{62.29} & 0.5066 & 
8 & \textcolor{red}{24.20} & 0.6539 & \textcolor{blue}{0.2100} & 0.7185 & 59.08 & 0.4655 \\

9 & 24.08 & 0.6485 & 0.2049 & \textcolor{red}{0.7327} & 62.10 & \textcolor{red}{0.5142} & 
9 & 24.18 & 0.6536 & 0.2108 & 0.7180 & 58.95 & 0.4648 \\

10 & 24.03 & 0.6477 & \textcolor{red}{0.2040} & \textcolor{blue}{0.7324} & \textcolor{blue}{62.14} & \textcolor{blue}{0.5126} & 
10 & 24.17 & 0.6534 & 0.2112 & 0.7141 & 58.79 & 0.4645 \\

11 & 24.05 & 0.6484 & \textcolor{blue}{0.2046} & 0.7320 & 62.12 & 0.5123 & 
11 & 24.16 & 0.6528 & 0.2117 & \textcolor{red}{0.7192} & 58.48 & 0.4613 \\

12 & 24.06 & 0.6485 & 0.2051 & 0.7320 & 61.33 & 0.5014 & 
12 & 24.16 & 0.6528 & 0.2118 & \textcolor{blue}{0.7191} & 58.68 & 0.4622 \\

\hline
% =========================================================
% Block 4: clip=6 (Left) vs clip=7 (Right)
% =========================================================
\multirow{2}{*}{$\tau_{\text{lpips}}$} & \multicolumn{6}{c|}{$\tau_{\text{clipiqa}}=6$} & \multirow{2}{*}{$\tau_{\text{lpips}}$} & \multicolumn{6}{c}{$\tau_{\text{clipiqa}}=7$} \\
 & PSNR & SSIM & LPIPS($\downarrow$) & CLIPIQA & MUSIQ & MANIQA &  & PSNR & SSIM & LPIPS($\downarrow$) & CLIPIQA & MUSIQ & MANIQA \\ \hline

% 左边从6开始，右边从7开始
6 & 24.18 & 0.6549 & 0.2108 & \textcolor{blue}{0.6946} & \textcolor{red}{58.96} & 0.4621 & 
 & & & & & & \\ 

7 & \textcolor{blue}{24.19} & \textcolor{red}{0.6557} & \textcolor{red}{0.2095} & 0.6931 & 58.85 & \textcolor{red}{0.4697} & 
\cellcolor{gray!25}7 & \cellcolor{gray!25}24.23 & \cellcolor{gray!25}\textcolor{red}{0.6563} & \cellcolor{gray!25}\textcolor{blue}{0.2102} & \cellcolor{gray!25}0.6651 & \cellcolor{gray!25}58.15 & \cellcolor{gray!25}\textcolor{blue}{0.4595} \\

\cellcolor{gray!25}8 & \cellcolor{gray!25}\textcolor{red}{24.20} & \cellcolor{gray!25}\textcolor{blue}{0.6556} & \cellcolor{gray!25}\textcolor{blue}{0.2106} & \cellcolor{gray!25}\textcolor{red}{0.6982} & \cellcolor{gray!25}58.85 & \cellcolor{gray!25}0.4625 & 
8 & \textcolor{red}{24.23} & \textcolor{blue}{0.6560} & \textcolor{red}{0.2097} & 0.6659 & 58.24 & 0.4535 \\

9 & 24.18 & 0.6550 & 0.2111 & 0.6928 & 58.92 & 0.4653 & 
9 & \textcolor{blue}{24.19} & 0.6555 & 0.2103 & 0.6667 & 58.31 & \textcolor{red}{0.4597} \\

10 & 24.17 & 0.6549 & 0.2117 & 0.6914 & 58.73 & \textcolor{blue}{0.4668} & 
10 & \textcolor{blue}{24.19} & 0.6552 & 0.2110 & 0.6664 & 58.25 & 0.4562 \\

11 & 24.18 & 0.6553 & 0.2116 & 0.6911 & 58.57 & 0.4591 & 
11 & 24.17 & 0.6551 & 0.2115 & \textcolor{blue}{0.6668} & \textcolor{blue}{58.41} & 0.4541 \\

12 & \textcolor{red}{24.20} & 0.6552 & 0.2118 & 0.6912 & \textcolor{blue}{58.94} & 0.4604 & 
12 & \textcolor{blue}{24.19} & 0.6549 & 0.2116 & \textcolor{red}{0.6681} & \textcolor{red}{58.95} & 0.4569 \\

\hline
% =========================================================
% Block 5: clip=8 (Left) vs clip=9 (Right)
% =========================================================
\multirow{2}{*}{$\tau_{\text{lpips}}$} & \multicolumn{6}{c|}{$\tau_{\text{clipiqa}}=8$} & \multirow{2}{*}{$\tau_{\text{lpips}}$} & \multicolumn{6}{c}{$\tau_{\text{clipiqa}}=9$} \\
 & PSNR & SSIM & LPIPS($\downarrow$) & CLIPIQA & MUSIQ & MANIQA &  & PSNR & SSIM & LPIPS($\downarrow$) & CLIPIQA & MUSIQ & MANIQA \\ \hline

% 左边从8开始，右边从9开始
8 & \textcolor{red}{24.05} & \textcolor{blue}{0.6474} & 0.2059 & 0.7215 & 59.60 & 0.5002 & 
 & & & & & & \\ 

9 & 24.04 & 0.6470 & 0.2037 & 0.7216 & 60.72 & \textcolor{blue}{0.5066} & 
\cellcolor{gray!25}9 & \cellcolor{gray!25}24.25 & \cellcolor{gray!25}\textcolor{blue}{0.6542} & \cellcolor{gray!25}\textcolor{red}{0.2087} & \cellcolor{gray!25}\textcolor{red}{0.6482} & \cellcolor{gray!25}\textcolor{red}{58.37} & \cellcolor{gray!25}\textcolor{blue}{0.4564} \\

10 & 24.03 & 0.6464 & \textcolor{blue}{0.2026} & 0.7256 & \textcolor{blue}{61.06} & 0.5062 & 
10 & \textcolor{red}{24.25} & \textcolor{red}{0.6544} & 0.2092 & 0.6474 & 58.20 & 0.4506 \\

\cellcolor{gray!25}11 & \cellcolor{gray!25}24.03 & \cellcolor{gray!25}0.6466 & \cellcolor{gray!25}0.2026 & \cellcolor{gray!25}\textcolor{blue}{0.7313} & \cellcolor{gray!25}\textcolor{red}{61.15} & \cellcolor{gray!25}\textcolor{red}{0.5067} & 
11 & 24.21 & 0.6536 & 0.2097 & \textcolor{blue}{0.6481} & 58.23 & \textcolor{red}{0.4577} \\

12 & 24.02 & \textcolor{red}{0.6478} & \textcolor{red}{0.2023} & \textcolor{red}{0.7377} & 60.89 & 0.5049 & 
12 & 24.22 & 0.6538 & 0.2092 & 0.6477 & 58.07 & 0.4554 \\

\hline
\end{tabular}%
}
\end{table*}
\section{Runtime \& Computational Complexity}
To assess the feasibility of IAFS in practical deployment scenarios, we conducted a systematic computational cost analysis, benchmarking our proposed method against existing inference-time scaling techniques. This section first outlines the experimental setup and testing protocols, followed by a quantitative comparison of runtime latency and peak GPU memory usage across different strategies.
\subsection{Experimental Setup}
 All performance benchmarks were conducted in a unified hardware environment featuring a single NVIDIA GeForce RTX 4090 GPU (24GB VRAM) paired with an Intel Xeon Gold 6248 CPU, implemented using the PyTorch 2.1 framework with CUDA 12.1 acceleration. To ensure reliable measurements, we recorded the average inference time for generating a single $256 \times 256$ resolution image and the peak GPU memory usage, with all reported results averaged over 50 consecutive runs following a warm-up phase to eliminate initialization jitter.
\subsection{Comparative Analysis}
Tab.~\ref{tab:time} details the computational costs and performance metrics for each method on the ImageNet dataset using the ResShift backbone, where comparative methods (BON~\cite{ma2025inference}, BS~\cite{li2025dynamic}, FK~\cite{singhal2025general}, KDS~\cite{hu2025kernel}) adopt $N=10$ candidate particles and employ CLIPIQA as the first-iteration reward with the same segmented scheduling in the next iteration. As shown, BON achieves the lowest latency due to its simple parallel sampling, whereas KDS incurs the highest cost because of expensive kernel density optimization, making it impractical for real-time use. Our IAFS strikes a balanced trade-off, and its frequency decomposition and adaptive steering introduce only modest overhead, reflected by the small time increase from $n=1$ to $n=2$. Memory usage remains around 6.00 GB for most search-based methods, indicating that storage of high-dimensional latent variables constitutes the primary bottleneck. Overall, IAFS demonstrates deployment-friendly efficiency, offering strong frequency-aware guidance within standard memory budgets and avoiding the extreme latency of optimization-heavy baselines, with future improvements potentially achievable via adaptive particle pruning to enhance real-time applicability.

\section{More Qualitative Results}
\begin{figure*}[t]
    \centering 
    \includegraphics[width=0.9\linewidth]{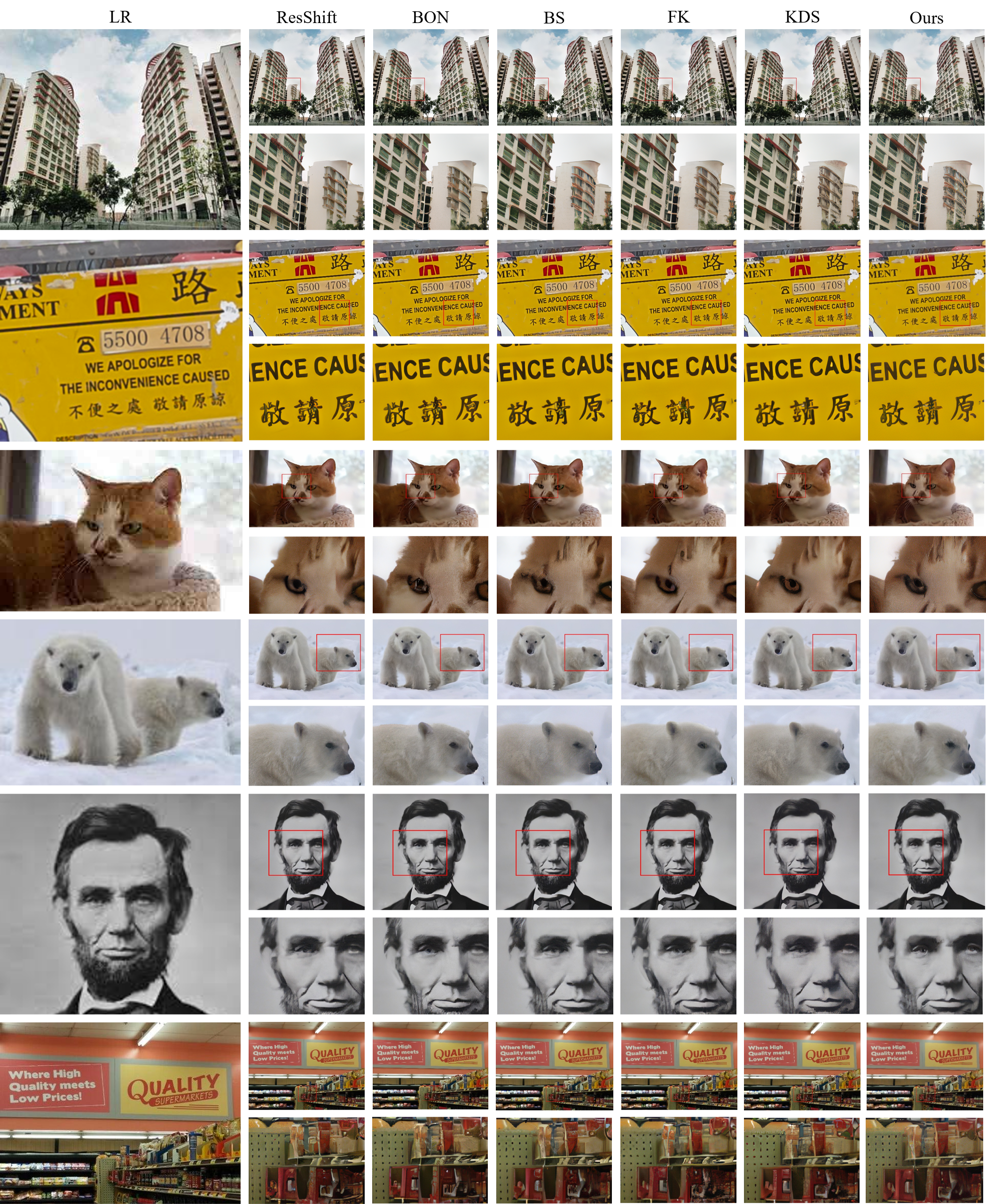} 
    \caption{Extended qualitative comparison on real-world images.} 
    \label{fig:real} % 标签
\end{figure*}

To further substantiate the robustness of our proposed IAFS framework, we provide an extended qualitative comparison on real-world images. We integrate various inference-time scaling strategies—specifically Best-of-$N$ (BON), Beam Search (BS), FK-Steering (FK), and Kernel Density Steering (KDS)—into the ResShift backbone and evaluate their performance against our method.
Fig.~\ref{fig:real} presents visual results across diverse scenarios, including architectural structures, text, animal fur, and human faces. While search-based strategies like BON, BS, and FK generally improve sharpness compared to the vanilla ResShift baseline, they often introduce semantic artifacts or lose structural fidelity. KDS relies on kernel density estimation to guide particles toward high-density regions, tends to sacrifice high-frequency texture for stability. These visual comparisons align with our quantitative findings, confirming that the adaptive frequency steering mechanism allows the model to selectively enhance high-frequency details while strictly adhering to the low-frequency structural constraints, thereby generating super-resolved images that are both visually pleasing and faithful to the input. The result further demonstrate that IAFS maintains consistent performance across diverse real-world distributions, highlighting its robustness and practical applicability.

\section{Potential Limitations}
While the proposed Iterative Diffusion Inference-Time Scaling with Adaptive Frequency Steering (IAFS) effectively mitigates the inherent conflict between perceptual enhancement and structural fidelity in super-resolution, the method still presents notable limitations. The primary bottleneck originates from the computational overhead introduced during inference. Specifically, IAFS requires sampling a pool of $N$ particles per timestep and executing $n$ refinement iterations to progressively adjust and stabilize the frequency components. As a result, the overall inference time increases approximately linearly with respect to both $N$ and $n$. Despite this limitation, the added computation is justified in applications demanding high-frequency consistency, fine-grained textures, and robust structural preservation, where IAFS demonstrably reduces over-smoothing and suppresses structural hallucination compared to existing approaches.


\end{document}